\documentclass[11pt]{article}

\usepackage[preprint]{acl}

\usepackage{times}
\usepackage{latexsym}

\usepackage[T1]{fontenc}

\usepackage[utf8]{inputenc}
\usepackage{algorithm}
\usepackage{graphicx}
\usepackage{booktabs}
\usepackage{multirow}
\usepackage{amsmath}
\usepackage{amssymb}
\usepackage{algorithm}
\usepackage{algpseudocode}
\usepackage{xcolor}
\usepackage{enumitem}
\usepackage{microtype}

\usepackage{inconsolata}

\usepackage{graphicx}

\title{Do Language Models Know Which Words Steer Their Reasoning?}

\author{
  \textbf{Yanbei Jiang\qquad Chao Lei\qquad Jey Han Lau}\\[0.2em]
  \textbf{Krista A. Ehinger\qquad Rena Gao Wei}\\[0.4em]
  \textnormal{University of Melbourne}
}

\begin{document}
\maketitle
\begin{abstract}
While chain-of-thought (CoT) prompting elicits intermediate reasoning steps from language models, it is unclear which words in the context actually steer the CoT. We identify \textit{Phantom Context Influence} (PCI): language models are
disproportionately steered by semantically irrelevant context words, even as
the same models deny those words' relevance when asked.
To localize PCI, we introduce Leave-One-Out Word JSD analysis, which removes
one context word at a time from the otherwise intact context to isolate its
marginal effect on the next-token distribution; words whose removal flips the
prediction (pivot words) are passed to the same LLM to self-judge whether it recognizes this predictive dependence.
Across four models and three benchmarks, language models
largely do \emph{not} know which words steer their reasoning, a mismatch we
call the Generation-Judgment Contradiction.
Inspired by this phenomenon, we propose Online KV Suppression: it scores every
prompt word's influence over the whole chain of thought in a single pass, then
zeroes the KV-cache entries of the highest-influence word the model itself
judges inert and redecodes.
This single-word intervention improves CoT
accuracy across four models and five reasoning benchmarks, by up to $8.9$
percentage points on math reasoning benchmark.
\end{abstract}

\section{Introduction}
\label{sec:introduction}

Chain-of-thought (CoT) prompting is now a standard approach for eliciting
structured reasoning from
LLMs~\citep{wei2022chain,kojima2022large,song2025introspect,ye2026faithfulness}.
However, each intermediate token is generated conditional on the entire
preceding context, whereas the contribution of individual context words to a
given prediction remains unclear. Consequently, next-token predictions may
depend substantially on context words that are semantically irrelevant to the
underlying problem. Such dependencies can alter intermediate predictions and
propagate through subsequent generation, thereby changing the resulting
reasoning trajectory without a corresponding semantic justification. This hidden dependence undermines interpretability and may
ultimately steer the model toward an incorrect answer.

\begin{figure}[t]
  \centering
  \includegraphics[width=0.45\textwidth]{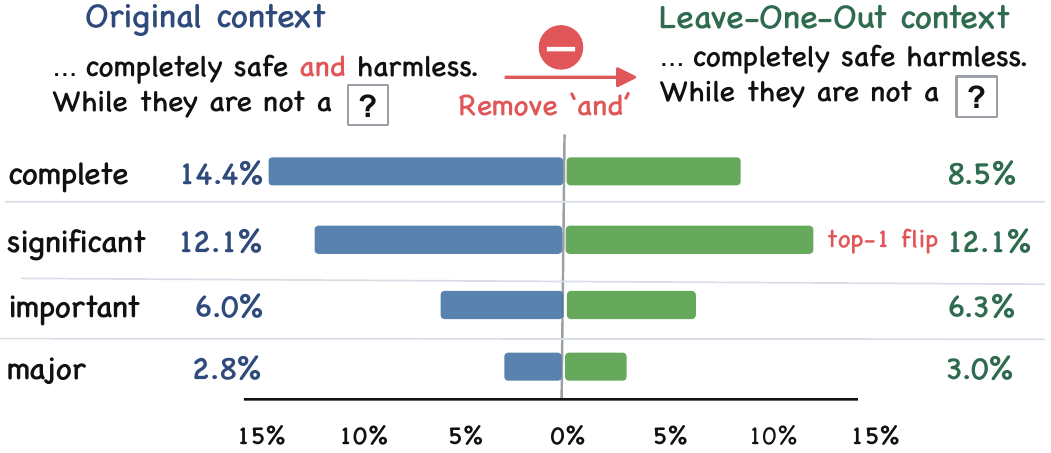}
  \caption{Illustration of leave-one-out analysis. Removing `and' from the original context shifts the next-token distribution and flips the top-1 prediction.}
  \vspace{-1.0em}
  \label{fig:leave_one_out}
\end{figure}

Figure~\ref{fig:leave_one_out} illustrates how a semantically irrelevant
context word can substantially alter the model's next-token distribution.
Given the original partial context, the model predicts \textit{complete} with
a probability of 14.4\%. Removing \textit{`and'} while holding all other
context words fixed reduces this probability to 8.5\%, whereas
\textit{significant} remains at 12.1\% and is consequently selected instead.
Although \textit{`and'} provides no domain-specific evidence for resolving the
factual question, its removal changes the model's top-1 prediction, indicating
that predictive influence can diverge from semantic relevance.

When asked to evaluate the effect of removing \textit{and}, however, the same
model predicts no change in its output. We call this {generation-judgment contradiction}
mismatch --- i.e., the discrepancy between model's
 sensitivity to certain context words during generation and their
subsequent classification of those words as irrelevant to the question --- \emph{Phantom Context Influence} (PCI):
context words that provide little information relevant to the question
nevertheless substantially alter the model's next-token distribution.

PCI is distinct from positional recency bias~\citep{liu2024lost}, which
characterizes aggregate effects of token position in long contexts, and from
chain-level CoT unfaithfulness~\citep{turpin2023language}, which concerns the
correspondence between a generated rationale and the model's final decision.
PCI instead operates at the level of individual next-token predictions,
identifying specific context words whose presence alters the predicted token
without corresponding semantic relevance.

To quantify PCI, we introduce \emph{Leave-One-Out Word JSD analysis}. 
For each
generated CoT word $W_t$, we compare the model's next-token distribution under
the full preceding context with the distribution obtained after removing the
$k$-th preceding context word, $W_{t-k}$. 
The Jensen--Shannon divergence between
the two distributions quantifies the influence of $W_{t-k}$.
We define $W_{t-k}$ as a \emph{pivot word} if its removal
changes the top-1 prediction, i.e., the argmax of the next-token distribution. 
We then use the same LLM to self-assess whether the model recognizes the
predictive dependence on each pivot word by asking whether removing
$W_{t-k}$ would change the prediction of the target word $W_t$. 
A pivot
word is identified as an instance of PCI if the model judges that its
removal would not change the prediction. Because exhaustive word-level
detection is computationally expensive, we further introduce a
response-level analysis that applies the same leave-one-out principle to
each prompt word and measures its influence on the complete generated
response, followed by an LLM Self-Judge assessment of whether the model
recognizes this response-level dependence. The response includes both CoT and final answer. Our experimental results show that, across the evaluated models and tasks,
models frequently judge that removing pivot words would not change
the corresponding prediction or response at both levels.


Based on this diagnosis, we propose \emph{Online KV Suppression}, an
inference-time intervention for mitigating PCI during CoT generation. Online
KV Suppression extends Leave-One-Out Word JSD analysis from individual
next-token predictions to the complete generated response. Specifically, it
evaluates each prompt word according to the change in the reasoning chain's
joint log-probability caused by removing that word, thereby capturing its
cumulative influence across the reasoning process. The method then identifies
PCI instances among the prompt words, ranks them by this leave-one-out change,
zeros the KV-cache entries associated with the highest-ranked instance, and
regenerates the response. This procedure requires only one leave-one-out
evaluation for each prompt word, rather than repeated PCI detection at every
decoding step, thereby avoiding the quadratic cost of token-level detection.

Online KV Suppression is entirely self-contained and requires no external models. Across four models and five benchmarks, it improves reasoning
accuracy by up to $8.9$ percentage points on arithmetic tasks. Notably, the
smallest model achieves the largest average gain ($+3.9$~pp), and the gains
remain positive for every model, with larger models gaining less.
Our contributions are:
\begin{enumerate}[noitemsep]
  \item We identify and systematically quantify \textbf{Phantom Context
  Influence} as a pervasive word-level phenomenon in LLM CoT generation,
  demonstrated across four models.
  \item We introduce \textbf{Leave-One-Out Word JSD} and \textbf{LLM Self-Judge}
  as a paired diagnostic framework for detecting PCI from output distributions
  and model self-assessments.
  \item We propose \textbf{Online KV Suppression}, an inference-time
  intervention guided by PCI diagnosis, and show that it improves CoT reasoning
  accuracy by suppressing spurious context dependencies during generation.
\end{enumerate}

\section{Related Work}


\paragraph{CoT faithfulness.}
\citet{turpin2023language} show that CoT explanations can be systematically
distorted by input formatting, producing plausible but misleading rationales.
\citet{lanham2023measuring, ye2026faithfulness} measure faithfulness via truncation and corruption
of reasoning chains, finding that later CoT tokens are often redundant for
the final answer.
\citet{arcuschin2025chain} further show that models construct superficially coherent
arguments to justify logically contradictory answers, even without artificially
inserted biases. \citet{lewislim2025cot} find that CoT influence and faithfulness are not always aligned, with instruction-tuned and reasoning models relying on CoT in
qualitatively different ways. The correctness of the individual inner CoT steps, however, remains little examined.


\paragraph{KV cache manipulation for long-context CoT.}
Longer CoT traces make KV-cache storage and attention increasingly costly,
motivating selective retention or eviction during inference. H$_2$O~\citep{zhang2023h2o}
keeps attention ``heavy hitters'', while StreamingLLM~\citep{xiao2024efficient}
preserves initial ``attention sinks'' plus a sliding window; both primarily
target efficiency and long-context stability. Recent work also questions the stability of token-importance estimates:
\citet{feng2025defensivekv} propose defensive aggregation to reduce worst-case
loss, and \citet{jiyang2026capkv} retain tokens via an information-theoretic
objective. Online KV Suppression uses the same primitive of selective KV editing,
but targets reasoning reliability rather than compression: it suppresses
dynamically identified spurious words that influence generation despite the model judging that their removal would not affect the output.


\paragraph{Local context dependence.}
Prior work shows that irrelevant context can disrupt LLM reasoning at the
problem level. \citet{shi2023large} find that inserted irrelevant sentences
reduce accuracy, and \citet{yang2025gsmdc} show in GSM-DC that this sensitivity
grows with distractor count and reasoning depth. Other work links irrelevant
context to hallucination through competing circuits~\citep{cheng2025stochastic}
and to shortened reasoning traces with reduced self-verification
~\citep{rodionov2026reasoning}.

These studies focus on sentence- or problem-level distraction. We instead
localize the effect to individual words inside the generated CoT and mitigate it
online through KV-cache suppression.

\section{Preliminary Analysis: Non-Monotonic Context Sensitivity}
\label{sec:preliminary}

Before defining PCI detection, we first show why a word-level diagnostic is
needed. During CoT generation, a model's next-token distribution is not shaped
evenly by the preceding words. Instead, removing specific context words can
create sharp distributional shifts.

\begin{figure}[t]
  \centering
  \includegraphics[width=0.65\columnwidth]{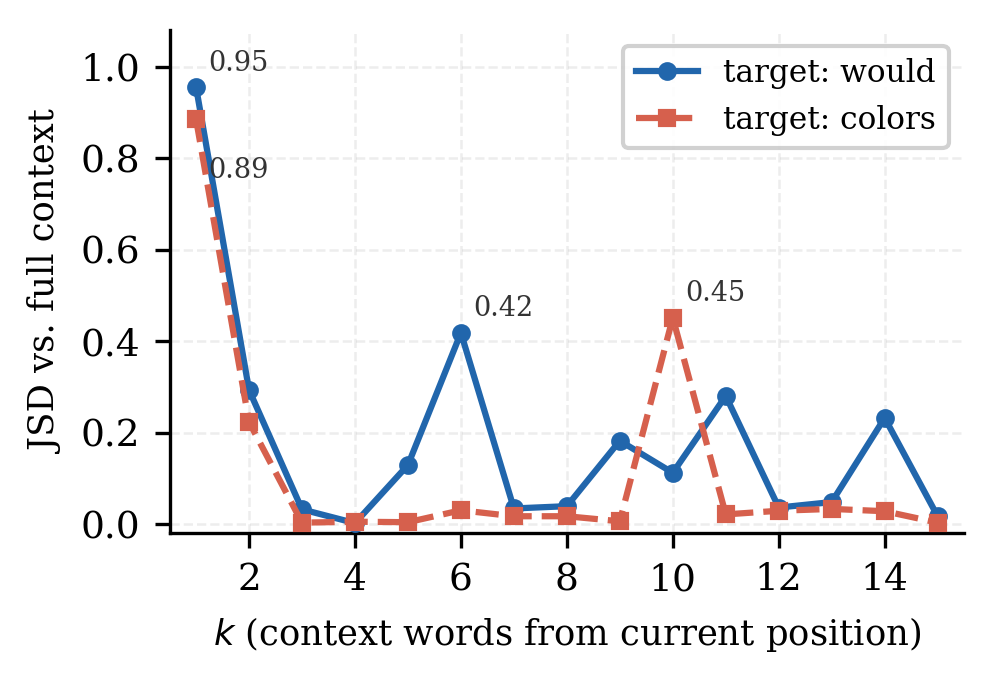}
  \caption{JSD between the full-context distribution and the distribution
  after removing the $k$-th previous word, for two target words during CoT
  generation (TruthfulQA, Qwen-3.5-4B).}
  \vspace{-1.0em}
  \label{fig:preliminary}
\end{figure}

Let $W_t$ be a target word generated during CoT.  $P_{\mathrm{full}}$ denotes
the model's next-token distribution conditioned on the prompt and the complete
generated context $W_1,\ldots,W_{t-1}$, and let $P_{\setminus k}$ denote the
distribution obtained after removing the $k$-th previous word $W_{t-k}$ while
keeping the rest of the context intact. We compare the two distributions:
\begin{equation}
  \widetilde{D}_k \;=\; \mathrm{JSD}\!\left(P_{\setminus k},\; P_{\mathrm{full}}\right).
  \label{eq:fullcontext_jsd}
\end{equation}
If each context word contributed gradually, $\widetilde{D}_k$
would remain small and decay smoothly with the distance $k$.
Figure~\ref{fig:preliminary} shows the opposite: the curves are highly
non-monotonic, with isolated spikes indicating that particular words have
disproportionate influence on the predicted distribution.

These patterns suggest that context sensitivity is not simply a matter of
recency or context length: the marginal influence of individual words is
highly non-uniform, and words far from the target can dominate the shift. 

\section{Detecting Phantom Context Influence}
\label{sec:adjk}

To analyze the non-monotonic context sensitivity of the preliminary analysis
quantitatively, we propose a framework for detecting Phantom Context
Influence. We now turn this observation into a systematic diagnosis. The goal
is to identify context words for which (i) removal from the otherwise intact
context demonstrably changes generation, yet (ii) the
model itself judges that the prediction would remain unchanged after the
removal. Following the preliminary
analysis, we first quantify the phenomenon at the \emph{word level}
(\S\ref{sec:loo_word}), testing each generated word against each of its
context words. Applied exhaustively, however, word-level detection requires
$O(T^2)$ leave-one-out forward evaluations for a chain of thought of length
$T$, which is impractical in practice. The framework therefore extends to the
\emph{response level} (\S\ref{sec:loo_resp}): a single leave-one-out pass
measures how removing each prompt word shifts the log-probability of the
entire generated response, reducing the cost to $O(T)$. We report the
Generation-Judgment Contradiction at both granularities
(\S\ref{sec:contradiction_exp}).

\subsection{Word-Level: Leave-One-Out Word JSD}
\label{sec:loo_word}

Let $x$ denote the original prompt and let
$\mathcal{W}=W_1,\ldots,W_T$ be the words generated during CoT. Each generated word may span one or more tokens. We group tokens into words
at whitespace and punctuation boundaries because the Self-Judge evaluates
word-level candidates, whereas predictive influence is measured over the
model's token distribution. For a target word $W_t$ consisting of one or more tokens, we use its first
token $w_t$ as the prediction target, since $w_t$ is predicted using only
the context preceding $W_t$.
For a target word $W_t$, we compare two contexts that differ by exactly
one word: the full preceding context and the same context with the $k$-th
previous word removed. For $k \geq 1$, define
\begin{align}
  C_{t} &= W_{1},\ldots,W_{t-1}, \\
  C^{\setminus k}_{t} &= W_{1},\ldots,W_{t-k-1},W_{t-k+1},\ldots,W_{t-1}.
\end{align}
Here $C^{\setminus k}_{t}$ is obtained by removing the single word $W_{t-k}$
from the full context $C_{t}$ while keeping all other words in place. Let
$P(\cdot \mid x, C)$ denote the model's next-token
distribution conditioned on the prompt $x$ and context $C$. We define
the Leave-One-Out (LOO) JSD at offset $k$ as
\begin{equation}
  D_{t,k}
  =
  \mathrm{JSD}\!\left(
    P(\cdot \mid x, C_{t}),
    P(\cdot \mid x, C^{\setminus k}_{t})
  \right).
  \label{eq:adjk_jsd}
\end{equation}
$D_{t,k}$ instantiates the removal-based measure of
Eq.~(\ref{eq:fullcontext_jsd}) for every target-candidate pair, 
isolating the
marginal effect of the single word $W_{t-k}$ on the prediction of $W_t$.

\paragraph{Pivot words.}
A large JSD alone is not enough: the distribution may change among tokens
unrelated to the word actually generated. We therefore focus on cases where
removing the word changes the argmax prediction and the flip involves $w_t$.
Let
\begin{align}
  \hat{w}^{\mathrm{full}}_{t} &= \operatorname*{arg\,max}_{w}
  P(w \mid x, C_{t}), \\
  \hat{w}^{\setminus k}_{t} &= \operatorname*{arg\,max}_{w}
  P(w \mid x, C^{\setminus k}_{t}).
\end{align}
We call $W_{t-k}$ a \textbf{pivot word} for target $W_t$ if
\begin{align}
  &\hat{w}^{\mathrm{full}}_{t} \neq \hat{w}^{\setminus k}_{t}, \notag \\
  &\hat{w}^{\mathrm{full}}_{t} = w_t
    \;\;\text{or}\;\;
    \hat{w}^{\setminus k}_{t} = w_t .
  \label{eq:transition}
\end{align}
The argmax condition ensures that the shift is directly tied to the generated
token rather than to an unrelated redistribution of probability mass.

\subsection{Response-Level: Leave-One-Out Influence}
\label{sec:loo_resp}

The word-level diagnostic asks whether one prediction depends on one context
word. Its response-level counterpart applies the same removal principle at
coarser granularity: does the \emph{whole} generated response depend on a
single prompt word? Let $X = (x_1, \ldots, x_n)$ denote the prompt tokens,
spanning $N$ words $W_1, \ldots, W_N$, and let
$\mathcal{R} = (s_1, \ldots, s_m)$ be the response generated greedily from 
$X$. For $k = 1, \ldots, N$, let $X^{\setminus k}$ denote $X$ with word $W_k$
removed and all other tokens kept in place. Under a context
$C \in \{X,\, X^{\setminus 1}, \ldots, X^{\setminus N}\}$, the response
average log-probability is the mean per-token log-probability of $\mathcal{R}$
under teacher-forcing:
\begin{equation}
  L(C) \;=\; \frac{1}{m}\sum_{j=1}^{m}
    \log P\!\left(s_j \;\middle|\; C,\; s_1,\ldots,s_{j-1}\right).
  \label{eq:chunk_logprob}
\end{equation}
We write $L_{\mathrm{full}} = L(X)$ and $L_{\setminus k} = L(X^{\setminus k})$.
The leave-one-out influence of prompt word $W_k$ on the response is
\begin{equation}
  \Delta_k \;=\; \bigl|L_{\setminus k} - L_{\mathrm{full}}\bigr|,
  \quad k = 1, \ldots, N.
  \label{eq:chunk_adjk}
\end{equation}
$\Delta_k$ is the \emph{absolute} change in mean log-probability per token:
we discard the sign, so a word whose removal makes the response more likely
counts as influential in the same way as one whose removal makes it less
likely, since either direction shows that the response is conditioned on
that word. Words with high $\Delta_k$ are the candidate pivot words.


\vspace{-0.1em}
\subsection{LLM Self-Judge}
\label{sec:judge}

Pivot words or response-level candidates are, 
by construction, words whose
removal demonstrably changes generation. The Self-Judge asks the same model
whether it is aware of this dependence. For a word-level candidate, the model
sees the problem and the complete CoT prefix with the candidate word marked in
place, and is asked whether deleting that word would change the next generated word. For a response-level candidate,
it sees the prompt with the candidate word marked and the full generated
response, and is asked whether the response depends on that word, i.e.,
whether removing it would change the response's claim or numbers. Both
questions elicit a binary verdict with in-context examples; the full templates
are given in Appendix~\ref{sec:prompt} and Appendix~\ref{sec:prompt_chunk}. A YES verdict indicates that the model predicts a change after removal (henceforth \textbf{load-bearing}),
whereas a NO verdict indicates that the model predicts no change (henceforth \textbf{inert}). To confirm that these verdicts
are not a model-specific artifact, we re-judge held-out samples with two
external models (GPT-5.6 and Gemini-3.1) and with human annotators: the
Self-Judge's verdicts agree substantially with all three
(Appendix~\ref{sec:cross_judge}). Moreover, replacing the Self-Judge with an
external judge inside the suppression pipeline yields equal or higher
accuracy and high agreement on the selected word
(Appendix~\ref{app:external_judge}), so our results do not hinge on the
generator judging itself.


\subsection{The Generation-Judgment Contradiction}
\label{sec:contradiction_exp}

A PCI instance is a pivot word that is causally influential yet judged inert:
\begin{equation}
  \mathrm{PCI}(t,k) \;=\;
\big[\text{Self-Judge}(W_{t-k}, w_t)=\text{NO}\big],
  \label{eq:scs}
\end{equation}
i.e., removing $W_{t-k}$ changes the prediction
(Eq.~\ref{eq:transition}), whereas the model judges that the prediction would remain unchanged. We call this the Generation-Judgment Contradiction. In generation, a word measurably shifts the model's prediction; in judgment, the same model declares that word irrelevant. The influence itself is not in question, it is established by measurement, so the only open variable is the model's verdict. And because the contradiction reproduces across models, the rate of these "irrelevant" verdicts gives a stable estimate of how often a model is steered by a word whose relevance it denies. 


\begin{figure}[t]
\centering
\includegraphics[width=0.499\textwidth]{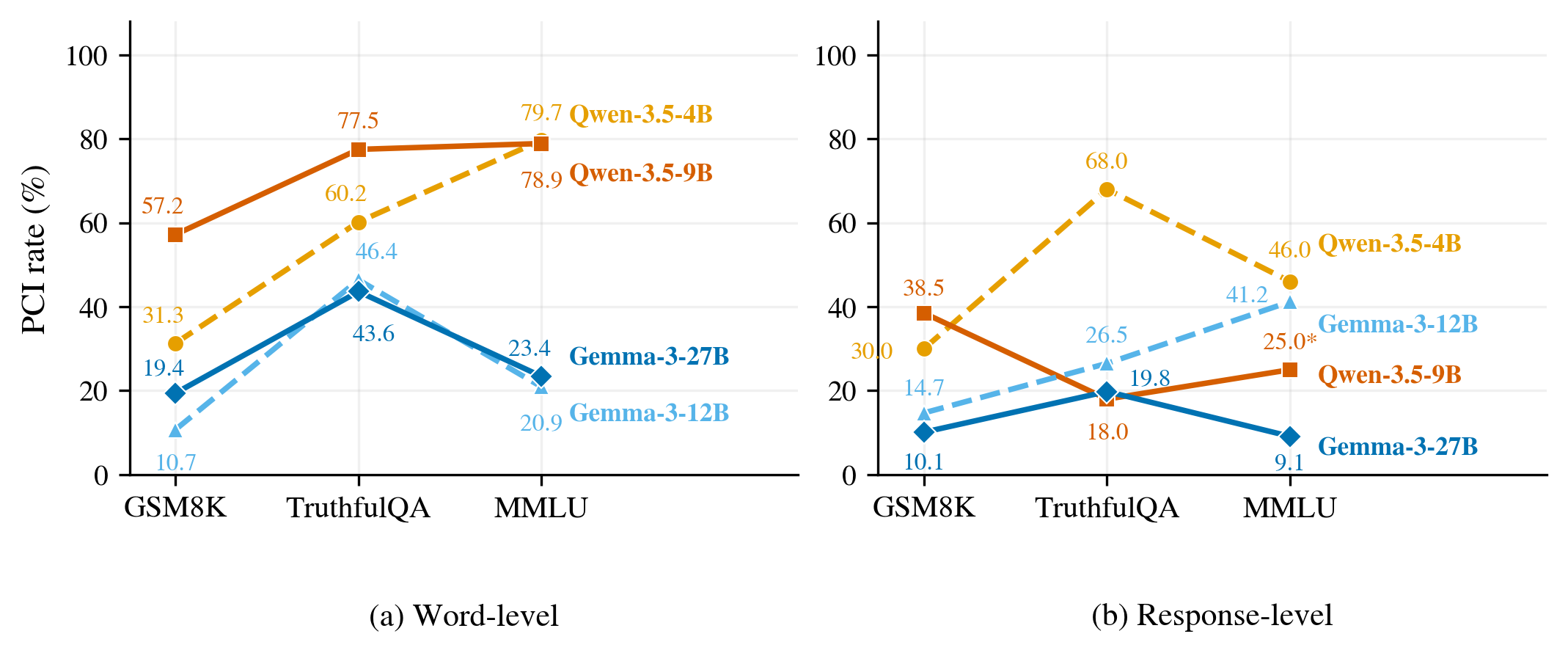}
\vspace{-1.5em}
\caption{Generation-Judgment Contradiction at both detection granularities.
(a) Word-level: fraction of pivot words (Eq.~\ref{eq:transition}) whose
measured influence is not recognized by the Self-Judge
(Eq.~\ref{eq:scs}), using 100 questions per dataset.
(b) Response-level: fraction of prompt words with substantial leave-one-out
influence (Eq.~\ref{eq:chunk_adjk})} 
\vspace{-1.0em}
\label{fig:contradiction}
\end{figure}

We run both diagnoses on the four models used throughout the
paper on the test set of
GSM8K~\citep{cobbe2021gsm8k}, TruthfulQA~\citep{lin2022truthfulqa}, and
MMLU~\citep{hendrycks2021mmlu}. For the response-level diagnosis we judge
every prompt word with $\Delta_k \geq 0.1$ nats/token ($\delta$ = 0.1), since only removals
that induce a substantial shift in the response are meaningful to analyze.

Figure~\ref{fig:contradiction} shows that the contradiction appears at both
granularities. At the word level (Fig.~\ref{fig:contradiction}a), the Qwen models exhibit the contradiction most
strongly (PCI rates
range from 31.3\% to 79.7\%), while the Gemma models are
more conservative (10.7--46.4\%) but still disown 
a substantial minority of
the words that steer their own predictions. The phenomenon does not shrink
with scale within a family: Qwen-3.5-9B exceeds the smaller 4B on GSM8K and
TruthfulQA and matches it on MMLU, and Gemma-3-27B exceeds Gemma-3-12B on
GSM8K and MMLU. At the response level (Fig.~\ref{fig:contradiction}b), rates
are lower, 9.1\% to 68.0\%, which is expected: a word whose removal shifts
the entire response is more often genuinely load-bearing. We also do an ablation study for different threshold $\delta$. Raising the candidate
threshold $\delta$ beyond the default $0.1$ shrinks the candidate set and
generally lowers the rate further, consistent with this skew
(Appendix~\ref{sec:threshold_sens}). The contradiction
nonetheless persists on every model--dataset pair, and the family ordering
(Qwen above Gemma) is consistent across the two granularities. The
contradiction is therefore not an artifact of any single detection choice:
whether influence is measured on one prediction or on the whole chain of
thought, every model disowns a meaningful share of the words that provably
steer its generation.
Human annotation supports these verdicts: on $150$ pivot words judged
irrelevant by the model ($50$ per dataset, Qwen-3.5-4B), human annotators
rate $87\%$ as indeed superfluous to the reasoning step
(Appendix~\ref{app:human_pci}).



\section{Suppressing Phantom Context at Inference Time}
\label{sec:suppress}

The Generation-Judgment Contradiction suggests a natural
intervention: a prompt word that is causally influential on the generated
response, yet whose influence the model itself denies, is a candidate for
suppression. The question is \emph{which} word to suppress: in principle any
prompt word or even several at once, could be targeted. The most intuitive
rule is to remove the single word with the highest influence, i.e., the
top-$\Delta_k$ candidate. We compare this and alternative selection rules
(influence-only selection, random candidates, words the judge deems
dependent, and multi-word variants) in an ablation study (Table~\ref{tab:ablation_yes_no}), and find that
suppressing the highest-influence word \emph{that the Self-Judge judges
inert} is the superior strategy: high influence alone does not imply
spuriousness, and the top-ranked candidates are often exactly the operands 
the response genuinely needs. Our method therefore composes the two signals:
use influence to \emph{rank}, use the judgment to \emph{filter}, and suppress
the single highest-influence word.

Because the response-level score (\S\ref{sec:loo_resp}) is computed for all
prompt words in one batched $O(T)$ pass, this design also stays
online-friendly. The inference cost is a constant factor of two to three over
a zero-shot decode (Appendix~\ref{app:latency}): where zero-shot performs one decoding pass, our method
performs two (the initial generation and a single redecode), plus one
batched scoring pass and a few Self-Judge queries, which are issued in
descending order of influence and stop at the first inert verdict. The total
cost thus still scales linearly with response length.

\subsection{Method}
\label{sec:method}

\paragraph{Ranking and judging.}
We rank the prompt
words by $\Delta_k$ in descending order and query the Self-Judge top-down with
the response-dependence question of \S\ref{sec:judge} (full template in
Appendix~\ref{sec:prompt_chunk}). An inert verdict marks the word as spurious:
the response does not rely on it.


\paragraph{Single-word selection.}
We suppress only the \emph{first} word judged inert in the $\Delta$-ranked
scan, the highest-influence prompt word the model itself judges non-essential.
Suppressing this single word keeps the intervention minimal; an ablation
that instead suppresses the top-3 or top-5 inert words (Table~\ref{tab:ablation_yes_no}) finds no gains compared to top-1.

\paragraph{KV suppression procedure.}

Although leave-one-out detection explicitly removes a word, direct deletion during intervention would (i) shift all subsequent token positions, confounding suppression with changes in positional encoding, and (ii) alter the input itself, so the model would no longer answer the original, intact question. We therefore preserve the token sequence and zero the selected word's key and value vectors in every layer of the prefilled cache. Zeroing a key sets the attention logit at that position to zero, so the position still receives a share of the softmax mass but, with its value zeroed, contributes nothing to the attention output; this is not identical to masking the position out of the normalization, and it leaves intact whatever information about the word was already routed into the representations of later prompt tokens during prefill. The operator thus removes the word's direct contribution while keeping positions and content unchanged, and the accuracy gains it yields are evidence of influence, not by themselves proof that the word is semantically irrelevant (Appendix~\ref{app:human_pci} reports human judgments). Detection, by contrast, only ranks and screens candidates, where deletion is the natural textual counterfactual, the same one the Self-Judge assesses. All suppression baselines use the identical position-preserving operator, so accuracy differences reflect which words are selected, not how suppression is performed.\footnote{Using KV suppression as the detection operator as well changes accuracy by at most $0.3$~pp and selects the same word on over $90\%$ of questions (Appendix~\ref{app:detection_operator}).}
Generation proceeds in one detect--redecode cycle:
\begin{enumerate}[noitemsep, leftmargin=*]
  \item \textbf{Generate.}
    Greedily decode the full response $\mathcal{R}$ from the prompt $X$.

  \item \textbf{Detect.}
    Compute $\Delta_k$ (Eqs.~\ref{eq:chunk_logprob}--\ref{eq:chunk_adjk})
    for every prompt word in a single batched pass.

  \item \textbf{Select.}
    Rank the remaining words by
    $\Delta_k$ in descending order, and query the Self-Judge top-down until
    the first word judged inert, $w^\star$. If no such word exists, return
    $\mathcal{R}$ unchanged.

  \item \textbf{Suppress.}
    Rebuild the KV cache $\mathcal{K}$ over the prompt $X$ and zero the entries
    at the token positions of $w^\star$:
    \begin{equation}
      \mathbf{K}_j \leftarrow \mathbf{0},\quad
      \mathbf{V}_j \leftarrow \mathbf{0},\quad
      j \in \mathrm{pos}\!\left(w^\star\right).
      \label{eq:suppression}
    \end{equation}

  \item \textbf{Redecode.}
    Regenerate the response greedily from the suppressed cache to EOS.
\end{enumerate}

\begin{table*}[t]
\centering
\small
\setlength{\tabcolsep}{3pt}
\begin{tabular}{llcccccc}
\toprule
Model & Method & GSM8K & GSM-Sym. & TruthfulQA & MMLU & AQUA-RAT & Avg. \\
\midrule
\multirow{5}{*}{Qwen-3.5-4B}
  & Zero-Shot      & 72.3 & 72.2 & 64.3 & 59.0 & 69.7 & 67.5 \\
  & High-Attn.     & 72.6 & 68.3 & 64.5 & 59.5 & 68.9 & 66.8 \\
  & Low-Attn.      & 75.2 & 71.8 & 64.7 & 59.9 & 71.7 & 68.7 \\
  & Stopword-Supp. & 75.8 & 72.6 & 64.2 & 58.7 & 69.7 & 68.2 \\
  & \textbf{Ours}  & \textbf{81.2} & \textbf{76.5} & \textbf{65.0} & \textbf{61.1} & \textbf{73.2} & \textbf{71.4} \\
\midrule
\multirow{5}{*}{Qwen-3.5-9B}
  & Zero-Shot      & 74.9 & 82.2 & 71.6 & 70.2 & 82.3 & 76.2 \\
  & High-Attn.     & 77.3 & 81.2 & 71.7 & 68.6 & 72.4 & 74.2 \\
  & Low-Attn.      & 75.2 & 82.5 & 72.0 & 69.8 & 81.1 & 76.1 \\
  & Stopword-Supp. & 78.4 & 79.6 & 71.8 & \textbf{70.4} & 79.1 & 75.9 \\
  & \textbf{Ours}  & \textbf{81.0} & \textbf{82.5} & \textbf{72.2} & 70.2 & \textbf{82.3} & \textbf{77.6} \\
\midrule
\multirow{5}{*}{Gemma-3-12B}
  & Zero-Shot      & 86.5 & 82.0 & 57.5 & 57.9 & 65.4 & 69.9 \\
  & High-Attn.     & 87.2 & 83.3 & 55.9 & 53.2 & 65.0 & 68.9 \\
  & Low-Attn.      & 86.1 & 82.3 & 56.8 & 58.5 & 66.7 & 70.1 \\
  & Stopword-Supp. & 86.6 & 81.5 & 57.1 & 58.4 & 65.7 & 69.9 \\
  & \textbf{Ours}  & \textbf{87.8} & \textbf{84.7} & \textbf{57.6} & \textbf{59.4} & \textbf{67.3} & \textbf{71.4} \\
\midrule
\multirow{5}{*}{Gemma-3-27B}
  & Zero-Shot      & 91.6 & 88.8 & 66.0 & 66.8 & 79.1 & 78.5 \\
  & High-Attn.     & 76.1 & 75.0 & 57.6 & 58.4 & 72.4 & 67.9 \\
  & Low-Attn.      & 91.7 & 87.3 & 63.7 & 64.7 & 78.7 & 77.2 \\
  & Stopword-Supp. & 86.8 & 82.3 & 64.3 & 65.0 & 77.6 & 75.2 \\
  & \textbf{Ours}  & \textbf{93.4} & \textbf{90.7} & \textbf{66.5} & \textbf{67.1} & \textbf{80.3} & \textbf{79.6} \\
\bottomrule
\end{tabular}
\caption{Accuracy (\%) on four models and five datasets.
\textbf{Ours} is Online KV Suppression (\S\ref{sec:method}); Zero-Shot,
High-Attn., Low-Attn., and Stopword-Supp.\ are baselines.
\textbf{Bold} = best method per model--dataset column.}
\vspace{-1.0em}
\label{tab:main_results}
\end{table*}

Suppression persists across the entire redecoded response rather than applying
only to the first token because $\Delta_k$ is defined over the
full response. Because only one prompt word is zeroed and the response is
regenerated in a single pass, the intervention is minimal and cannot cascade.
Appendix Algorithm~\ref{alg:suppress} formalizes the complete procedure.

\subsection{Main Results}

\paragraph{Experimental settings.}
We evaluate on five benchmarks:
GSM8K~\citep{cobbe2021gsm8k},
GSM-Symbolic~\citep{mirzadeh2024gsm},
TruthfulQA~\citep{lin2022truthfulqa},
MMLU~\citep{hendrycks2021mmlu} (five subjects),
and AQUA-RAT~\citep{ling2017aqua}. Four target models are evaluated:
Qwen-3.5-4B, Qwen-3.5-9B, Gemma-3-12B-it, and Gemma-3-27B-it.
Accuracy is measured by exact numerical match (GSM8K, GSM-Symbolic),
multiple-choice answer matching (MMLU, AQUA-RAT), and
an LLM judge (TruthfulQA, GPT-5.5). 


\paragraph{Baselines.}
We include two suppression baselines that bypass the leave-one-out detection and Self-Judge pipeline.
\textbf{High-Attn.}\ suppresses the top-$5\%$ of prompt tokens ranked by the
attention the generated response pays to them;
\textbf{Low-Attn.}\ suppresses the bottom-$5\%$ from the same candidate pool and
\textbf{Stopword-Supp.}\ suppresses a fixed list of English function words (prepositions and common discourse markers) present in the prompt. All reuse the same single-pass suppress-and-redecode mechanism as our method but replace the Self-Judge selection with a fixed rule, requiring no per-instance detection.

\paragraph{Results Analysis} Online KV Suppression improves accuracy over the zero-shot baseline on nearly every model--dataset cell (Table~\ref{tab:main_results}) and is the best of the four methods on all but one cell (Qwen-3.5-9B MMLU), where Stopword-Supp.\ edges ahead by $0.2$~pp.
Average gains are largest for the smallest model and smaller for the larger ones, $+3.9$~pp (Qwen-3.5-4B), $+1.4$~pp (Qwen-3.5-9B), $+1.5$~pp (Gemma-3-12B), and $+1.1$~pp (Gemma-3-27B), a pattern consistent with weaker models being more susceptible to spurious context.
The largest gains appear on arithmetic reasoning, plausibly because a single spurious prompt word can derail a multi-step computation. Consistent with this, nearly half of the first divergences between zero-shot and suppressed responses on GSM8K involve numbers, logic, or entities, and $23.0\%$ of intervened questions change their final answer, with corrections outnumbering regressions by $8.9$~pp (Appendix~\ref{app:local_vs_response}).
In relative terms, Qwen-3.5-4B reduces its GSM-Symbolic error by $15.5\%$ and Gemma-3-27B its GSM8K error by $21.4\%$, so the gains are most meaningful for arithmetic reasoning and for smaller models. 
High-Attn.\ hurts on average, suggesting that high-attention tokens are often semantically load-bearing, so that suppressing them by rank alone removes necessary context alongside spurious words. Pivot words do receive $1.7$--$4.1\times$ the average context attention (Appendix~\ref{app:attn}), so attention flags them but cannot separate spurious from necessary influence.
To probe whether these comparisons are an artifact of a single greedy decode, Appendix~\ref{app:stability} re-evaluates every cell with five temperature-sampled runs with the mean and standard deviation reported; 
the relative ordering of methods is generally preserved within this variation.
Finally, word clouds of the suppressed words (Appendix~\ref{app:wordclouds})
show that they are predominantly connectives, pronouns, determiners, and
negations.
\vspace{-0.2em}
\section{Discussion}
\vspace{-0.2em}
\paragraph{Ablation: suppressing necessary vs.\ spurious context.}

To probe whether the Self-Judge's verdicts carry real signal, we compare seven suppression-target settings on GSM8K and AQUA-RAT for both models (Table~\ref{tab:ablation_yes_no}): (1) no suppression (\emph{none}); (2) \emph{Influence-only}, suppressing the single highest-influence ($\Delta_k$) word with no Self-Judge filter; (3) \emph{Random}, suppressing one randomly selected candidate word; (4) \emph{Load-bearing}, suppressing the highest-influence word the judge marks as load-bearing; and (5)--(7) \emph{Ours (top-$K$)}, suppressing the $K\in\{1,3,5\}$ highest-influence words judged inert.
Suppressing inert words consistently improves accuracy. Influence-only is by far the most damaging condition, most plausibly because the top-ranked word is often a numeric operand or other necessary token; since Influence-only and Ours share the same influence ranking, we read this gap as mainly reflecting the value of the Self-Judge filter that Ours applies on top of it.
Load-bearing lowers accuracy in every setting, as expected if the judge's dependence verdicts identify genuinely necessary words. Furthermore, extending Ours from one suppressed word to three adds no gains for the two models. Two further controls that keep the judge but ignore the ranking, suppressing a random non-top inert word or the lowest-influence inert word, gain only $+1.5$ and $+0.7$~pp on average versus $+3.9$~pp for Ours (Appendix~\ref{app:inert_controls}).

\begin{table}[t]
\centering
\small
\setlength{\tabcolsep}{4pt}
\resizebox{0.8\columnwidth}{!}{%
\begin{tabular}{llcc}
\toprule
Model & Target & GSM8K & AQUA \\
\midrule
\multirow{7}{*}{Qwen-3.5-4B}
  & none            & 72.3 & 69.7 \\
  & Influence-only  & 44.2 & 60.6 \\
  & Random          & 73.5 & 68.5 \\
  & Load-bearing & 71.7 & 64.3 \\
  & Ours (top-1)    & \textbf{81.2} & \textbf{73.2} \\
  & Ours (top-3)    & 81.1 & 72.6 \\
  & Ours (top-5)    & \textbf{81.2} & 72.6 \\
\midrule
\multirow{7}{*}{Gemma-3-12B}
  & none            & 86.5 & 65.4 \\
  & Influence-only  & 67.9 & 60.2 \\
  & Random          & 84.2 & 65.8 \\
  & Load-bearing & 82.0 & 65.0 \\
  & Ours (top-1)    & \textbf{87.8} & \textbf{67.3} \\
  & Ours (top-3)    & \textbf{87.8} & \textbf{67.3} \\
  & Ours (top-5)    & \textbf{87.8} & \textbf{67.3} \\
\bottomrule
\end{tabular}
}
\caption{Accuracy (\%) of seven suppression-target settings on two models.}
\vspace{-1.0em}
\label{tab:ablation_yes_no}
\end{table}

\paragraph{Ablation: response-level vs.\ chunk-level suppression.}
A finer-grained alternative (chunk-level
suppression) segments the response at punctuation boundaries and, at each
chunk, runs leave-one-out detection over all preceding context (prompt \emph{and}
already-generated tokens), suppressing the highest-influence non-numeric
inert word for that chunk. Chunk-level thus targets locally generated words, whereas
response-level targets prompt words globally. Table~\ref{tab:full_vs_chunk}
compares the two under the identical judge prompt and numeric guard.
Response-level is better or tied on average for both models and notably stronger on
arithmetic, plausibly because the answer there often hinges on a small number of
influential prompt words.
Chunk-level is competitive on knowledge tasks (TruthfulQA, MMLU), possibly because the relevant spurious word there is more often generated mid-reasoning than given in the prompt.

\begin{table}[t]
\centering
\small
\setlength{\tabcolsep}{3pt}
\resizebox{\columnwidth}{!}{%
\begin{tabular}{llcccccc}
\toprule
Model & Variant & GSM8K & GSM-S. & TQA & MMLU & AQUA & Avg. \\
\midrule
\multirow{2}{*}{Qwen-3.5-4B}
  & Chunk    & 75.4 & 72.7 & \textbf{65.7} & 60.6 & 70.5 & 69.0 \\
  & Response & \textbf{81.2} & \textbf{76.5} & 65.0 & \textbf{61.1} & \textbf{73.2} & \textbf{71.4} \\
\midrule
\multirow{2}{*}{Gemma-3-12B}
  & Chunk    & 86.9 & 84.0 & \textbf{61.3} & \textbf{60.3} & 62.6 & 71.0 \\
  & Response & \textbf{87.8} & \textbf{84.7} & 57.6 & 59.4 & \textbf{67.3} & \textbf{71.4} \\
\bottomrule
\end{tabular}
}
\caption{Response-level (Ours) vs.\ chunk-level suppression, same judge prompt
and numeric guard.}
\vspace{-1.0em}
\label{tab:full_vs_chunk}
\end{table}

\paragraph{Upper bound: best-of-$k$ spurious-word suppression.}
The default method suppresses only the single highest-influence spurious word.
To probe how much single-word suppression could recover if it always chose the
right word, we form an oracle upper bound: from the JSD ranking we take the
top-$k$ words the Self-Judge labels spurious (numeric operands excluded),
suppress each one independently, redecode, and count a question correct if anyone is correct (pass@$k$). We compare against a matched zero-shot
pass@$k$ baseline that draws $k$ sampled generations. For every question the $k$ candidates are the first $k$ entries of one fixed influence-ranked list of inert words, so the pass@5 and pass@10 candidate sets are nested and both are evaluated on the identical question set (a question with fewer than $k$ inert words simply has fewer attempts). From Table~\ref{tab:upperbound}, at $k{=}5$ our suppression outperforms zero-shot pass@$k$ on all four settings, whereas at $k{=}10$ it does so only for the two Gemma-3-12B settings, as the diversity of sampled generations grows faster with $k$ than the pool of distinct inert words. Because each candidate comes from suppressing a specific, model-identified
spurious word, the upper bound is interpretable: the answers it recovers are attributable to
removing individual spurious dependencies.

\begin{table}[t]
\centering
\small
\setlength{\tabcolsep}{5pt}
\resizebox{\columnwidth}{!}{%
\begin{tabular}{llcccc}
\toprule
& & \multicolumn{2}{c}{GSM8K} & \multicolumn{2}{c}{AQUA-RAT} \\
\cmidrule(lr){3-4}\cmidrule(lr){5-6}
Model & Method & @5 & @10 & @5 & @10 \\
\midrule
\multirow{2}{*}{Qwen-3.5-4B}
  & Zero-shot pass@$k$ & 90.8 & \textbf{93.9} & 87.8 & \textbf{92.1} \\
  & Ours pass@$k$   & \textbf{92.1} & 92.4 & \textbf{90.2} & 90.9 \\
\midrule
\multirow{2}{*}{Gemma-3-12B}
  & Zero-shot pass@$k$ & 92.8 & 94.0 & 83.1 & 85.8 \\
  & Ours pass@$k$   & \textbf{94.1} & \textbf{94.4} & \textbf{84.3} & \textbf{86.2} \\
\bottomrule
\end{tabular}
}
\caption{Upper bound results for pass@$k$.}
\vspace{-1.0em}
\label{tab:upperbound}
\end{table}

\section{Conclusion}

The Leave-One-Out Word JSD framework localizes PCI at word granularity from
output distributions alone, and the LLM Self-Judge turns the contradiction into
a practical detection signal.
Online KV Suppression operationalizes this and demonstrates that the
contradiction is actionable: scoring each prompt word over the entire chain of
thought in a single pass and zeroing the KV entries of the one
highest-influence word the model itself judges spurious improves CoT accuracy
across four models and five benchmarks.


\paragraph{Limitations.}
Our evidence covers four open-weight models from two families (Qwen-3.5 and Gemma-3); the method requires white-box access to the KV cache, so we could not test closed frontier models and do not claim that the findings extend to them. Because only the single highest-influence spurious word is suppressed, errors stemming from multiple spurious dependencies or non-contextual reasoning mistakes are left uncorrected, and the Self-Judge's binary verdict may be too coarse where relevance is graded rather than categorical. KV zeroing removes a word's direct attention contribution but is not equivalent to attention masking or to deleting the word, so gains under suppression do not by themselves establish semantic irrelevance; and the response-level Self-Judge prompt contains explicit guidance that favors labeling function words inert, which we did not test against a less leading formulation. 


\bibliography{custom}

@inproceedings{cheng2025stochastic,
  title={Stochastic Chameleons: Irrelevant Context Hallucinations Reveal
         Class-Based ({Mis})Generalization in {LLM}s},
  author={Cheng, Ziling and Cao, Meng and Rondeau, Marc-Antoine
          and Cheung, Jackie Chi Kit},
  booktitle={Proceedings of the 63rd Annual Meeting of the Association
             for Computational Linguistics},
  year={2025},
  publisher={Association for Computational Linguistics}
}

@article{rodionov2026reasoning,
  title={Reasoning Shift: How Context Silently Shortens {LLM} Reasoning},
  author={Rodionov, Gleb},
  journal={arXiv preprint arXiv:2604.01161},
  year={2026}
}

@inproceedings{yang2025gsmdc,
  title={How Is {LLM} Reasoning Distracted by Irrelevant Context?
         {A}n Analysis Using a Controlled Benchmark},
  author={Yang, Minglai and Huang, Ethan and Zhang, Liang
          and Surdeanu, Mihai and Wang, William Yang and Pan, Liangming},
  booktitle={Proceedings of the 2025 Conference on Empirical Methods
             in Natural Language Processing},
  pages={13329--13347},
  year={2025},
  publisher={Association for Computational Linguistics}
}

@inproceedings{shi2023large,
  title={Large Language Models Can Be Easily Distracted by Irrelevant Context},
  author={Shi, Freda and Chen, Xinyun and Misra, Kanishka and Scales, Nathan
          and Dohan, David and Chi, Ed H. and Sch{\"a}rli, Nathanael and Zhou, Denny},
  booktitle={Proceedings of the 40th International Conference on Machine Learning},
  pages={31210--31227},
  year={2023},
  publisher={PMLR}
}

@inproceedings{wei2022chain,
  title={Chain-of-thought prompting elicits reasoning in large language models},
  author={Wei, Jason and Wang, Xuezhi and Schuurmans, Dale and Bosma, Maarten
          and Ichter, Brian and Xia, Fei and Chi, Ed and Le, Quoc V and Zhou, Denny},
  booktitle={Advances in Neural Information Processing Systems},
  volume={35},
  pages={24824--24837},
  year={2022}
}

@inproceedings{kojima2022large,
  title={Large language models are zero-shot reasoners},
  author={Kojima, Takeshi and Gu, Shixiang Shane and Reid, Machel
          and Matsuo, Yutaka and Iwasawa, Yusuke},
  booktitle={Advances in Neural Information Processing Systems},
  volume={35},
  pages={22199--22213},
  year={2022}
}

@article{arcuschin2025chain,
  title={Chain-of-Thought Reasoning In The Wild Is Not Always Faithful},
  author={Arcuschin, Iv{\'a}n and Janiak, Jett and Krzyzanowski, Robert
          and Rajamanoharan, Senthooran and Nanda, Neel and Conmy, Arthur},
  journal={arXiv preprint arXiv:2503.08679},
  year={2025}
}

@article{lewislim2025cot,
  title={Analysing Chain of Thought Dynamics: Active Guidance or Unfaithful
         Post-hoc Rationalisation?},
  author={Lewis-Lim, Samuel and Tan, Xingwei and Zhao, Zhixue and Aletras, Nikolaos},
  journal={arXiv preprint arXiv:2508.19827},
  year={2025}
}

@article{ye2026faithfulness,
  title={Mechanistic Evidence for Faithfulness Decay in Chain-of-Thought Reasoning},
  author={Ye, Donald and Loffgren, Max and Kotadia, Om and Wong, Linus},
  journal={arXiv preprint arXiv:2602.11201},
  year={2026}
}

@inproceedings{turpin2023language,
  title={Language models don't always say what they think:
         Unfaithful explanations in chain-of-thought prompting},
  author={Turpin, Miles and Michael, Julian and Perez, Ethan and Bowman, Samuel},
  booktitle={Advances in Neural Information Processing Systems},
  volume={36},
  year={2023}
}

@article{lanham2023measuring,
  title={Measuring faithfulness in chain-of-thought reasoning},
  author={Lanham, Tamera and Chen, Anna and Radhakrishnan, Ansh and Steiner, Benoit
          and Denison, Carson and Hernandez, Danny and Li, Dustin and Durmus, Esin
          and Hubinger, Evan and Kernion, Jackson and others},
  journal={arXiv preprint arXiv:2307.13702},
  year={2023}
}

@article{feng2025defensivekv,
  title={Taming the Fragility of {KV} Cache Eviction in {LLM} Inference},
  author={Feng, Yuan and Guo, Haoyu and Lv, JunLin and Zhou, S. Kevin and Xie, Xike},
  journal={arXiv preprint arXiv:2510.13334},
  year={2025}
}

@article{jiyang2026capkv,
  title={Rethinking {KV} Cache Eviction via a Unified Information-Theoretic Objective},
  author={Yang, Jiaming and Tang, Chenwei and Zhen, Liangli and Lv, Jiancheng},
  journal={arXiv preprint arXiv:2604.25975},
  year={2026}
}

@inproceedings{zhang2023h2o,
  title={{H$_2$O}: Heavy-hitter oracle for efficient generative inference
         of large language models},
  author={Zhang, Zhenyu and Sheng, Ying and Zhou, Tianyi and Chen, Tianlong
          and Zheng, Lianmin and Cai, Ruisi and Song, Zhao and Tian, Yuandong
          and R{\'e}, Christopher and Barrett, Clark and others},
  booktitle={Advances in Neural Information Processing Systems},
  volume={36},
  year={2023}
}

@inproceedings{xiao2024efficient,
  title={Efficient streaming language models with attention sinks},
  author={Xiao, Guangxuan and Tian, Yuandong and Chen, Beidi
          and Han, Song and Lewis, Mike},
  booktitle={International Conference on Learning Representations},
  year={2024}
}

@article{liu2024lost,
  title={Lost in the middle: How language models use long contexts},
  author={Liu, Nelson F and Lin, Kevin and Hewitt, John and Paranjape, Ashwin
          and Bevilacqua, Michele and Petroni, Fabio and Liang, Percy},
  journal={Transactions of the Association for Computational Linguistics},
  volume={12},
  pages={157--173},
  year={2024}
}

@article{song2025introspect,
  title={Language Models Fail to Introspect About Their Knowledge of Language},
  author={Song, Siyuan and Hu, Jennifer and Mahowald, Kyle},
  journal={arXiv preprint arXiv:2503.07513},
  year={2025}
}

@inproceedings{lin2022truthfulqa,
  title={{TruthfulQA}: Measuring how models mimic human falsehoods},
  author={Lin, Stephanie and Hilton, Jacob and Evans, Owain},
  booktitle={Proceedings of the 60th Annual Meeting of the Association
             for Computational Linguistics (Volume 1: Long Papers)},
  pages={3214--3252},
  year={2022}
}

@article{mirzadeh2024gsm,
  title={{GSM-Symbolic}: Understanding the limitations of mathematical reasoning
         in large language models},
  author={Mirzadeh, Iman and Alizadeh, Keivan and Shahrokhi, Hooman
          and Tuzel, Oncel and Bengio, Samy and Farajtabar, Mehrdad},
  journal={arXiv preprint arXiv:2410.05229},
  year={2024}
}

@article{cobbe2021gsm8k,
  title={Training Verifiers to Solve Math Word Problems},
  author={Cobbe, Karl and Kosaraju, Vineet and Bavarian, Mohammad
          and Chen, Mark and Jun, Heewoo and Kaiser, Lukasz
          and Plappert, Matthias and Tworek, Jerry and Hilton, Jacob
          and Nakano, Reiichiro and Hesse, Christopher and Schulman, John},
  journal={arXiv preprint arXiv:2110.14168},
  year={2021}
}

@article{hendrycks2021mmlu,
  title={Measuring Massive Multitask Language Understanding},
  author={Hendrycks, Dan and Burns, Collin and Basart, Steven
          and Zou, Andy and Mazeika, Mantas and Song, Dawn and Steinhardt, Jacob},
  journal={Proceedings of the International Conference on Learning Representations},
  year={2021}
}

@inproceedings{ling2017aqua,
  title={Program Induction by Rationale Generation: Learning to Solve and Explain Algebraic Word Problems},
  author={Ling, Wang and Yogatama, Dani and Dyer, Chris and Blunsom, Phil},
  booktitle={Proceedings of the 55th Annual Meeting of the Association for Computational Linguistics},
  pages={158--167},
  year={2017}
}

\appendix

\section{Stability of the Main Results}
\label{app:stability}
Table~\ref{tab:main_results} reports a single greedy decode per cell. To quantify
run-to-run stability, we re-evaluate every model--dataset--method cell with five
independent temperature-sampled runs ($\tau{=}1.0$, top-$p{=}1.0$) on a fixed
$300$-question subset per dataset (all $254$ for AQUA-RAT), holding the suppressed
words fixed to those of the main experiment (Ours: the Self-Judge's selected word;
Attn-/Stopword-Supp.: their deterministic rules) so that only the sampling varies.
This isolates the effect of decoding noise on the reported accuracies; it does
not re-run detection and judging, which are deterministic under greedy
generation and temperature-zero judging, so the table should be read as
decoding stability rather than as a repeated run of the full pipeline.
Table~\ref{tab:stability} reports mean\,$\pm$\,std over the five runs. Because
decoding is stochastic here rather than greedy, the center values differ from
Table~\ref{tab:main_results}; the standard deviations are small (mostly
$\le 2$~pp) and the relative ordering of the four methods is preserved to within
this run-to-run variation, indicating the effects reported in the main table are
stable rather than artifacts of a single decode.

\begin{table*}[t]
\centering
\small
\setlength{\tabcolsep}{5pt}
\begin{tabular}{llccccc}
\toprule
Model & Method & GSM8K & GSM-Sym. & TruthfulQA & MMLU & AQUA-RAT \\
\midrule
\multirow{4}{*}{Qwen-3.5-4B}
  & Zero-Shot & $74.1\,{\scriptstyle\pm1.0}$ & $72.8\,{\scriptstyle\pm1.6}$ & $64.6\,{\scriptstyle\pm1.1}$ & $60.4\,{\scriptstyle\pm1.7}$ & $70.2\,{\scriptstyle\pm1.2}$ \\
  & High-Attn. & $71.8\,{\scriptstyle\pm1.3}$ & $67.9\,{\scriptstyle\pm0.9}$ & $63.2\,{\scriptstyle\pm1.5}$ & $59.0\,{\scriptstyle\pm1.6}$ & $71.2\,{\scriptstyle\pm2.3}$ \\
      & Low-Attn. & $75.6\,{\scriptstyle\pm0.8}$ & $72.3\,{\scriptstyle\pm1.4}$ & $64.2\,{\scriptstyle\pm1.1}$ & $60.6\,{\scriptstyle\pm1.6}$ & $71.3\,{\scriptstyle\pm0.9}$ \\
  & Stopword-Supp. & $77.5\,{\scriptstyle\pm1.7}$ & $74.4\,{\scriptstyle\pm2.0}$ & $64.0\,{\scriptstyle\pm1.3}$ & $57.7\,{\scriptstyle\pm1.1}$ & $69.8\,{\scriptstyle\pm2.2}$ \\
  & \textbf{Ours} & $79.6\,{\scriptstyle\pm2.3}$ & $77.4\,{\scriptstyle\pm1.2}$ & $65.1\,{\scriptstyle\pm1.3}$ & $60.9\,{\scriptstyle\pm1.8}$ & $71.7\,{\scriptstyle\pm2.5}$ \\
\midrule
\multirow{4}{*}{Qwen-3.5-9B}
  & Zero-Shot & $73.3\,{\scriptstyle\pm1.5}$ & $80.8\,{\scriptstyle\pm1.8}$ & $71.2\,{\scriptstyle\pm1.5}$ & $70.5\,{\scriptstyle\pm1.4}$ & $81.4\,{\scriptstyle\pm1.4}$ \\
  & High-Attn. & $77.8\,{\scriptstyle\pm1.1}$ & $80.1\,{\scriptstyle\pm1.6}$ & $71.3\,{\scriptstyle\pm2.5}$ & $67.2\,{\scriptstyle\pm1.5}$ & $72.6\,{\scriptstyle\pm1.1}$ \\
      & Low-Attn. & $74.5\,{\scriptstyle\pm1.6}$ & $83.1\,{\scriptstyle\pm1.3}$ & $71.6\,{\scriptstyle\pm1.1}$ & $70.1\,{\scriptstyle\pm0.8}$ & $80.7\,{\scriptstyle\pm0.7}$ \\
  & Stopword-Supp. & $78.5\,{\scriptstyle\pm0.9}$ & $79.6\,{\scriptstyle\pm1.1}$ & $71.5\,{\scriptstyle\pm1.1}$ & $71.6\,{\scriptstyle\pm1.7}$ & $77.6\,{\scriptstyle\pm1.6}$ \\
  & \textbf{Ours} & $80.0\,{\scriptstyle\pm2.3}$ & $81.1\,{\scriptstyle\pm1.5}$ & $71.1\,{\scriptstyle\pm1.3}$ & $69.5\,{\scriptstyle\pm0.8}$ & $83.3\,{\scriptstyle\pm1.5}$ \\
\midrule
\multirow{4}{*}{Gemma-3-12B}
  & Zero-Shot & $84.2\,{\scriptstyle\pm2.3}$ & $82.0\,{\scriptstyle\pm0.7}$ & $57.4\,{\scriptstyle\pm1.1}$ & $57.8\,{\scriptstyle\pm1.9}$ & $65.2\,{\scriptstyle\pm1.9}$ \\
  & High-Attn. & $86.8\,{\scriptstyle\pm1.3}$ & $84.1\,{\scriptstyle\pm1.3}$ & $56.3\,{\scriptstyle\pm1.3}$ & $52.6\,{\scriptstyle\pm1.1}$ & $63.1\,{\scriptstyle\pm2.4}$ \\
    & Low-Attn. & $85.7\,{\scriptstyle\pm1.2}$ & $82.7\,{\scriptstyle\pm0.8}$ & $56.9\,{\scriptstyle\pm0.4}$ & $58.9\,{\scriptstyle\pm0.9}$ & $67.4\,{\scriptstyle\pm1.4}$ \\
  & Stopword-Supp. & $87.0\,{\scriptstyle\pm0.9}$ & $81.3\,{\scriptstyle\pm0.9}$ & $56.5\,{\scriptstyle\pm0.7}$ & $57.4\,{\scriptstyle\pm2.1}$ & $66.9\,{\scriptstyle\pm1.8}$ \\
  & \textbf{Ours} & $87.7\,{\scriptstyle\pm1.1}$ & $84.0\,{\scriptstyle\pm1.1}$ & $58.1\,{\scriptstyle\pm0.9}$ & $58.6\,{\scriptstyle\pm1.7}$ & $69.2\,{\scriptstyle\pm2.0}$ \\
\midrule
\multirow{4}{*}{Gemma-3-27B}
  & Zero-Shot & $92.3\,{\scriptstyle\pm1.0}$ & $87.9\,{\scriptstyle\pm1.2}$ & $65.9\,{\scriptstyle\pm0.7}$ & $66.1\,{\scriptstyle\pm1.1}$ & $79.7\,{\scriptstyle\pm0.9}$ \\
  & High-Attn. & $76.6\,{\scriptstyle\pm1.0}$ & $75.7\,{\scriptstyle\pm0.9}$ & $57.6\,{\scriptstyle\pm0.6}$ & $58.6\,{\scriptstyle\pm1.1}$ & $72.7\,{\scriptstyle\pm0.6}$ \\
      & Low-Attn. & $91.2\,{\scriptstyle\pm1.3}$ & $87.8\,{\scriptstyle\pm1.1}$ & $64.4\,{\scriptstyle\pm1.8}$ & $65.3\,{\scriptstyle\pm2.1}$ & $78.4\,{\scriptstyle\pm0.7}$ \\
  & Stopword-Supp. & $87.7\,{\scriptstyle\pm1.2}$ & $82.6\,{\scriptstyle\pm0.3}$ & $65.4\,{\scriptstyle\pm1.1}$ & $63.8\,{\scriptstyle\pm1.5}$ & $76.6\,{\scriptstyle\pm2.3}$ \\
  & \textbf{Ours} & $93.1\,{\scriptstyle\pm0.7}$ & $90.4\,{\scriptstyle\pm0.4}$ & $67.0\,{\scriptstyle\pm0.7}$ & $67.0\,{\scriptstyle\pm0.8}$ & $80.8\,{\scriptstyle\pm0.7}$ \\
\bottomrule
\end{tabular}
\caption{Run-to-run stability of Table~\ref{tab:main_results}. Each cell is
mean\,$\pm$\,std over five temperature-sampled runs ($\tau{=}1.0$, top-$p{=}1.0$) with the suppressed words
held fixed. Decoding is stochastic here (vs.\ greedy in the main table), so the
center values differ from Table~\ref{tab:main_results}.}
\label{tab:stability}
\end{table*}

\section{Word-Level Self-Judge Prompt Template}
\label{sec:prompt}

Used in \S\ref{sec:judge} to query the model about each pivot word.
The two Self-Judge prompts target different constructs: this word-level
prompt asks whether deleting the marked word would change the \emph{next
generated word}, whereas the response-level prompt
(Appendix~\ref{sec:prompt_chunk}) asks whether the \emph{whole response}
depends on the marked prompt word. They serve the two granularities of
\S\ref{sec:adjk} and \S\ref{sec:method} respectively and are not
interchangeable: a word can be inert for one next-token prediction yet
load-bearing for the response, and vice versa.
Fields: \texttt{\{transition\_word\}} = candidate word $W_{t-k}$;
\texttt{\{question\}} = the original problem;
\texttt{\{cot\_prefix\}} = the complete CoT prefix $W_1,\ldots,W_{t-1}$ with
the candidate word marked in place as \texttt{[[$W_{t-k}$]]};
\texttt{\{target\_word\}} = generated word $w_t$.

\begin{quote}
\small\ttfamily
Problem: ``\{question\}''\\
\\
Reasoning so far (the marked word is in [[ ]]):\\
``\{cot\_prefix\}''\\
\\
The next word generated after this reasoning is ``\{target\_word\}''.\\
\\
Question: If the marked word ``\{transition\_word\}'' were deleted (keeping
everything else the same), would the next word after the reasoning CHANGE?\\
\\
Answer YES if the next word would likely change (it would no longer be
``\{target\_word\}'').\\
Answer NO if the next word would likely stay the same (it would still be
``\{target\_word\}'').\\
\\
Reply with a single word: YES or NO
\end{quote}

%

\section{Self-Judge Prompt Template}
\label{sec:prompt_chunk}

Used in \S\ref{sec:method} to classify each prompt pivot word.
Fields: \texttt{\{transition\_word\}} = candidate prompt word $W_k$;
\texttt{\{question\_text\}} = the prompt with $W_k$ marked by \texttt{[[\,]]};
\texttt{\{cot\_text\}} = the generated response.
The judge is framed around \emph{dependence}, and returns a one-line reason
followed by a binary verdict. The template contains explicit guidance
(numbers and named quantities are almost always load-bearing; common function
words are almost always filler) and four examples. This guidance biases the
judge toward labeling function words inert, and the words our method
suppresses are correspondingly dominated by connectives, pronouns, and
determiners (Appendix~\ref{app:wordclouds}). We did not evaluate a less
leading formulation; the external-judge results
(Appendix~\ref{app:external_judge}) show that the gains do not depend on this
particular judge, but a neutral prompt remains to be tested.

\begin{quote}
\small\ttfamily
Decide whether the response is load-bearing on the marked word in the problem.\\
\\
Problem (target word marked [[ ]]): ``\{question\_text\}''\\
Generated response: ``\{cot\_text\}''\\
Target word: ``\{transition\_word\}''\\
\\
Does the response DEPEND on the target word --- does it use that word's value,
quantity, or specific meaning to reach its claim or numbers?\\
- YES: the response references or computes with the target word (removing it
would break or change the response). A number, quantity, or named result is
almost always YES.\\
- NO: the target word is incidental filler. Common words such as ``the'', ``a'',
``and'', ``so'', ``then'', ``he'', ``this'', ``let'', ``step'' are almost always
filler.\\
\\
Examples:\\
\phantom{ }target ``[[3]]'' in ``Tom has 3 apples, buys 2 more'': YES\\
\phantom{ }target ``[[\$40]]'' in ``a book costs \$40'': YES\\
\phantom{ }target ``[[Well,]]'' in ``Well, Tom has 3 apples'': NO\\
\phantom{ }target ``[[the]]'' in ``he sells the eggs'': NO\\
\\
Reply in the exact format:\\
Reason: at most 8 words\\
Answer: YES or NO
\end{quote}

\section{Threshold Sensitivity of the Response-Level PCI Rate}
\label{sec:threshold_sens}

Table~\ref{tab:threshold_sens} reports the response-level PCI rate as the
candidate threshold $\delta$ is raised from the default $0.1$ to $0.2$,
$0.3$, and $0.5$ nats/token, computed over the same judged words as
Figure~\ref{fig:contradiction}b (a word counts toward threshold $\delta$ if
$\Delta_k \geq \delta$). Candidate counts shrink rapidly with $\delta$, and
rates generally decline---consistent with the observation that the
highest-influence words skew load-bearing---while remaining substantial for
the Qwen models on arithmetic and knowledge tasks; cells with no remaining
candidates are marked ``--'', and rates over fewer than ten candidates
should be read with caution.

\begin{table*}[h]
\centering
\small
\setlength{\tabcolsep}{6pt}
\begin{tabular}{llcccc}
\toprule
\textbf{Model} & \textbf{Dataset} &
$\boldsymbol{\delta=0.1}$ & $\boldsymbol{\delta=0.2}$ &
$\boldsymbol{\delta=0.3}$ & $\boldsymbol{\delta=0.5}$ \\
\midrule
\multirow{3}{*}{Qwen-3.5-4B}
  & GSM8K      & 30.0 (293) & 26.7 (120) & 17.9 (39) & 20.0 (5) \\
  & TruthfulQA & 68.0 (269) & 48.7 (115) & 40.0 (55) & 29.4 (17) \\
  & MMLU       & 46.0 (137) & 40.0 (30)  & 30.0 (10) & 0.0 (3) \\
\midrule
\multirow{3}{*}{Qwen-3.5-9B}
  & GSM8K      & 38.5 (174) & 43.0 (93)  & 43.5 (69) & 44.2 (43) \\
  & TruthfulQA & 18.0 (50)  & 10.0 (20)  & 9.1 (11)  & 0.0 (5) \\
  & MMLU       & 25.0 (8)   & --         & --        & -- \\
\midrule
\multirow{3}{*}{Gemma-3-12B}
  & GSM8K      & 14.7 (102) & 0.0 (5)    & 0.0 (1)   & -- \\
  & TruthfulQA & 26.5 (226) & 8.0 (75)   & 6.5 (31)  & 20.0 (5) \\
  & MMLU       & 41.2 (119) & 35.7 (14)  & 0.0 (1)   & -- \\
\midrule
\multirow{3}{*}{Gemma-3-27B}
  & GSM8K      & 10.1 (158) & 3.0 (33)   & 0.0 (2)   & -- \\
  & TruthfulQA & 19.8 (308) & 9.5 (137)  & 1.4 (73)  & 0.0 (24) \\
  & MMLU       & 9.1 (154)  & 12.1 (33)  & 0.0 (8)   & 0.0 (4) \\
\bottomrule
\end{tabular}
\caption{Response-level PCI rate (\%) at increasing candidate thresholds
$\delta$ (nats/token); the number of candidates is given in parentheses.
``--'' marks cells with no candidates above the threshold.}
\label{tab:threshold_sens}
\end{table*}

\section{Human Validation of PCI Instances}
\label{app:human_pci}

To verify that the pivot words the model disowns are genuinely
superfluous, we sampled $150$ word-level PCI instances from Qwen-3.5-4B
($50$ per dataset, one per question, seed $42$): each is a pivot word whose
removal demonstrably changes the next-word prediction yet which the
Self-Judge marked irrelevant. Annotators saw the problem, the CoT prefix
with the pivot word marked, and the generated next word---the same view
given to the judge---and labeled whether the marked word is superfluous
for that reasoning step (items shuffled and blinded). Three co-authors served
as annotators, each labeling a disjoint set of $50$ items, so every item
received a single label and no inter-annotator agreement is reported;
annotators saw only the blinded items, with dataset and question identity
withheld until labeling was complete.

\begin{table}[h]
\centering
\small
\setlength{\tabcolsep}{5pt}
\resizebox{\columnwidth}{!}{%
\begin{tabular}{lccc}
\toprule
\textbf{Dataset} & \textbf{Items} & \textbf{Superfluous (\%)} &
\textbf{Not superfluous (\%)} \\
\midrule
GSM8K      & 50  & 84 & 16 \\
TruthfulQA & 50  & 90 & 10 \\
MMLU       & 50  & 86 & 14 \\
\midrule
Overall    & 150 & 87 & 13 \\
\bottomrule
\end{tabular}}
\caption{Human validation of word-level PCI instances (Qwen-3.5-4B).
Fraction of model-disowned pivot words that human annotators also judge
superfluous for the marked reasoning step.}
\label{tab:human_pci}
\end{table}

\section{Attention Evidence for PCI}
\label{app:attn}
\begin{figure}[t]
  \centering
  \includegraphics[width=\columnwidth]{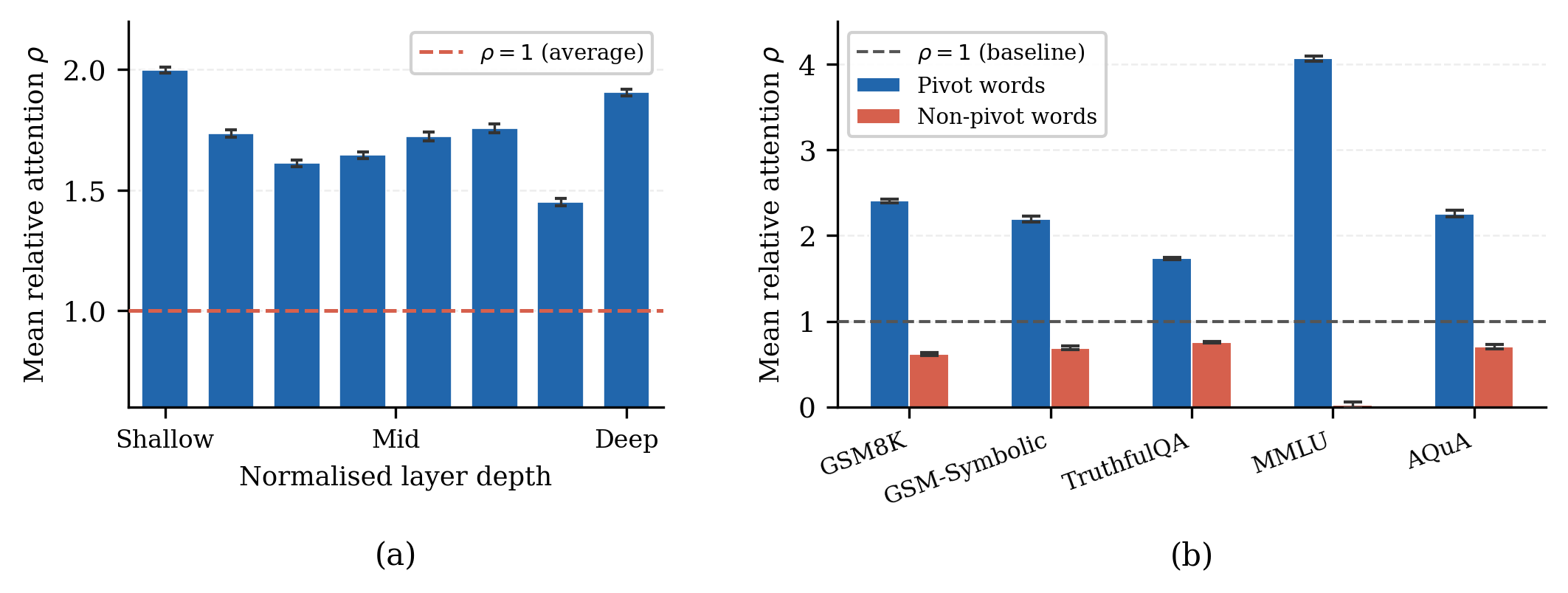}
  \caption{Attention evidence for PCI in Qwen-3.5-4B. (a) Mean relative
  attention $\rho$ of pivot words per sampled layer.
  (b) Pivot words (blue) vs.\ non-pivot words (red) in mean relative
  attention $\rho$, averaged over sampled layers.}
  \label{fig:attn}
\end{figure}

To corroborate the JSD-based diagnosis mechanistically, we compute the
relative attention weight that pivot words receive at the forward pass
for $W_t$:
\begin{equation}
  \rho(W_{t-k}) \;=\; \frac{\mathrm{attn}(q_{t} \to W_{t-k})}
  {\frac{1}{|\mathcal{P}|}\sum_{j \in \mathcal{P}} \mathrm{attn}(q_t \to W_j)},
\end{equation}
where $q_t$ is the query position of $W_t$ and $\mathcal{P}$ is the set of
all preceding context positions. 
Figure~\ref{fig:attn}(a) shows the per-layer mean $\rho$ on TruthfulQA:
pivot words receive well above their share of attention at every depth
($\rho \geq 1.45$), peaking at the shallowest layer ($\rho \approx 2.0$) and
remaining high through the deepest layer ($\rho \approx 1.9$).
Figure~\ref{fig:attn}(b) contrasts pivot words with
non-pivot words across five benchmarks for Qwen-3.5-4B, averaged over all
sampled layers: pivot words attract $1.7$--$4.1\times$ the average context
attention on every dataset (up to $\rho{=}4.1$ on MMLU),
while non-pivot words fall far below the $\rho{=}1$ baseline ($\rho \leq 0.75$).
Notably, a naive baseline that suppresses the highest-attention tokens
regardless of JSD behaviour performs poorly (High-Attn., Table~\ref{tab:main_results}),
demonstrating that elevated $\rho$ alone is insufficient to identify
spurious influence. Leave-one-out detection is necessary to pinpoint
which high-attention words distort the prediction.


\section{Word Clouds of Suppressed Words}
\label{app:wordclouds}

\begin{figure*}[t]
\centering
\begin{minipage}{0.49\textwidth}\centering
\includegraphics[width=\linewidth]{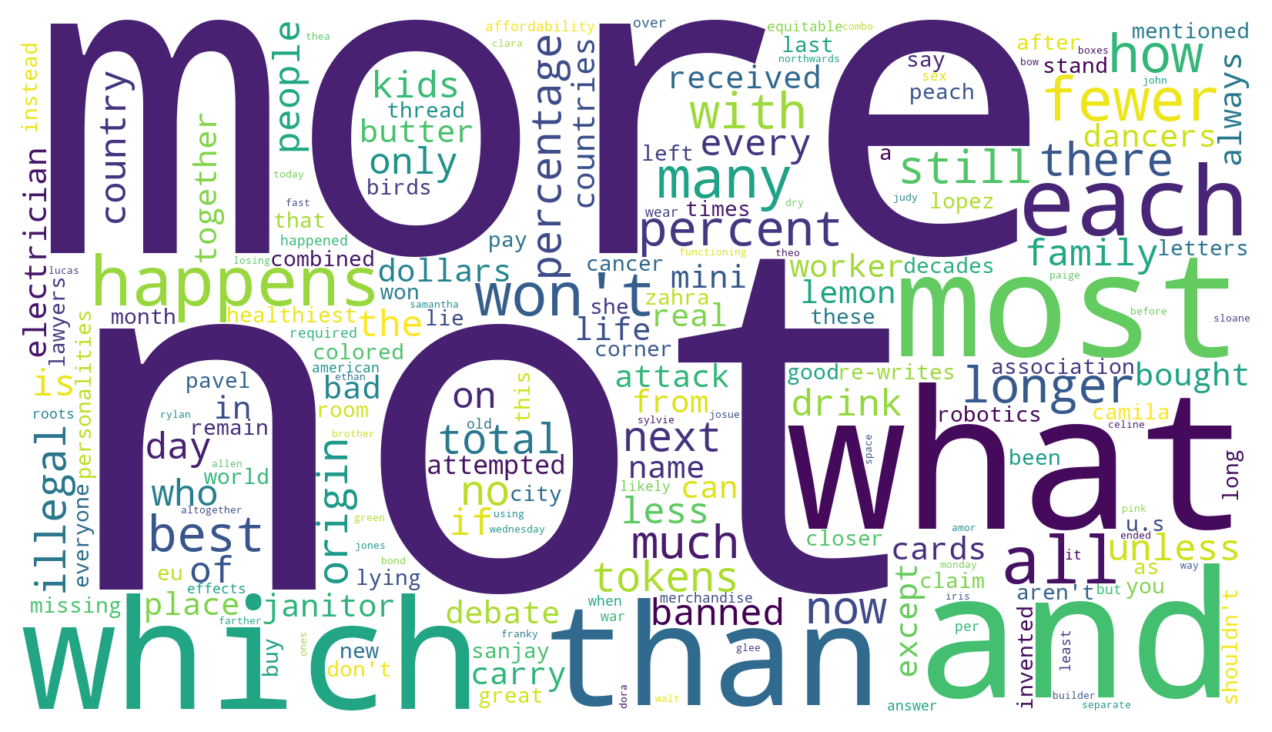}\\
{\small (a) Qwen-3.5-4B}
\end{minipage}\hfill
\begin{minipage}{0.49\textwidth}\centering
\includegraphics[width=\linewidth]{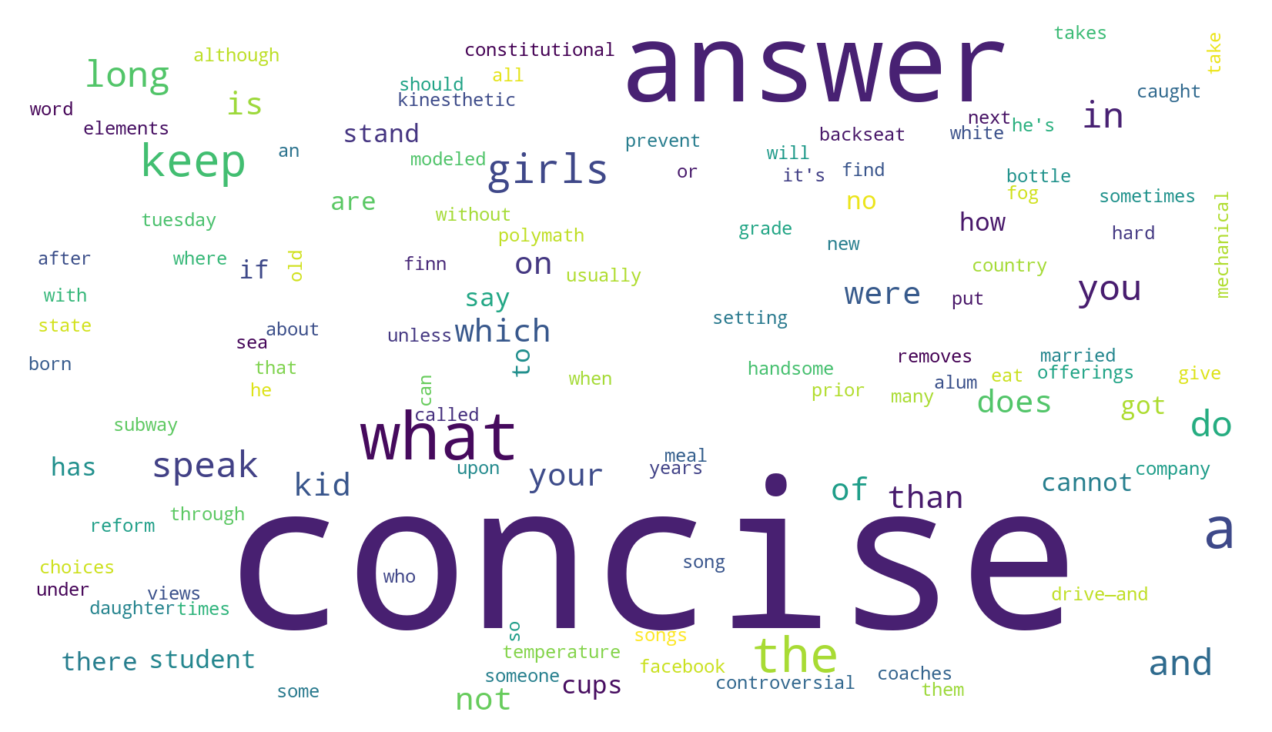}\\
{\small (b) Qwen-3.5-9B}
\end{minipage}\\[8pt]
\begin{minipage}{0.49\textwidth}\centering
\includegraphics[width=\linewidth]{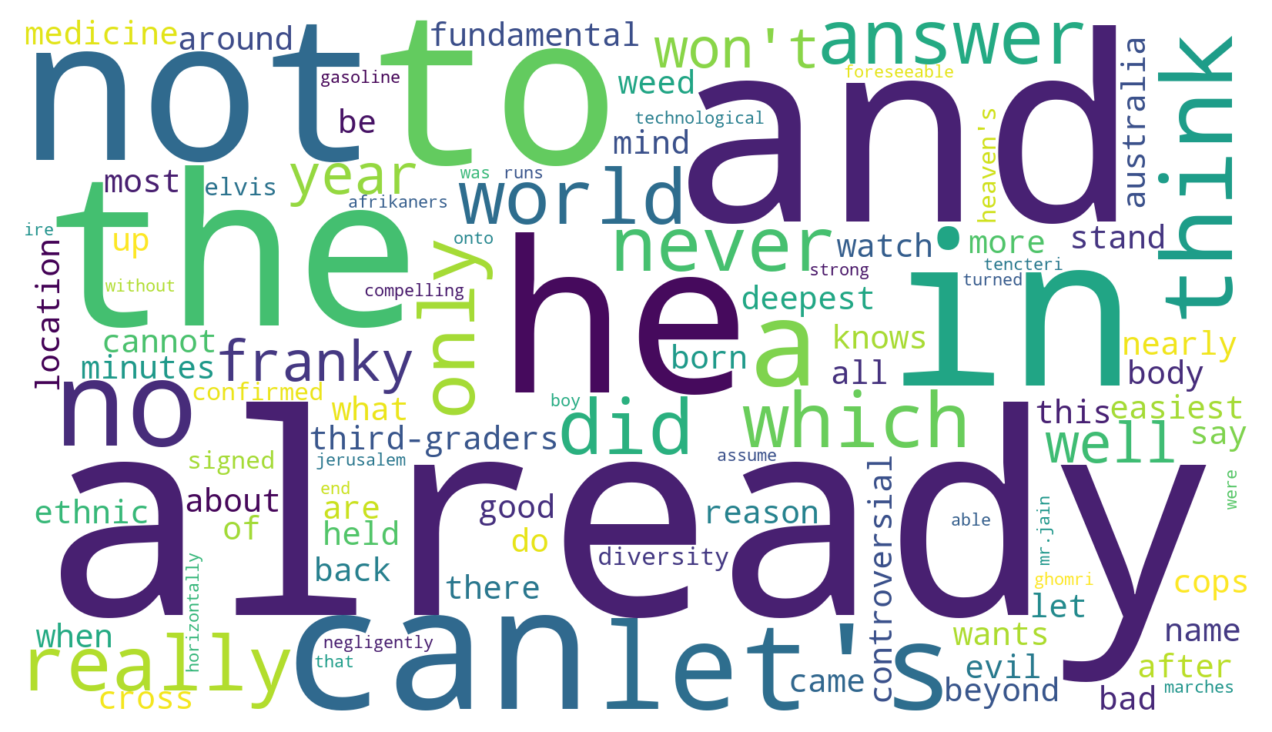}\\
{\small (c) Gemma-3-12B}
\end{minipage}\hfill
\begin{minipage}{0.49\textwidth}\centering
\includegraphics[width=\linewidth]{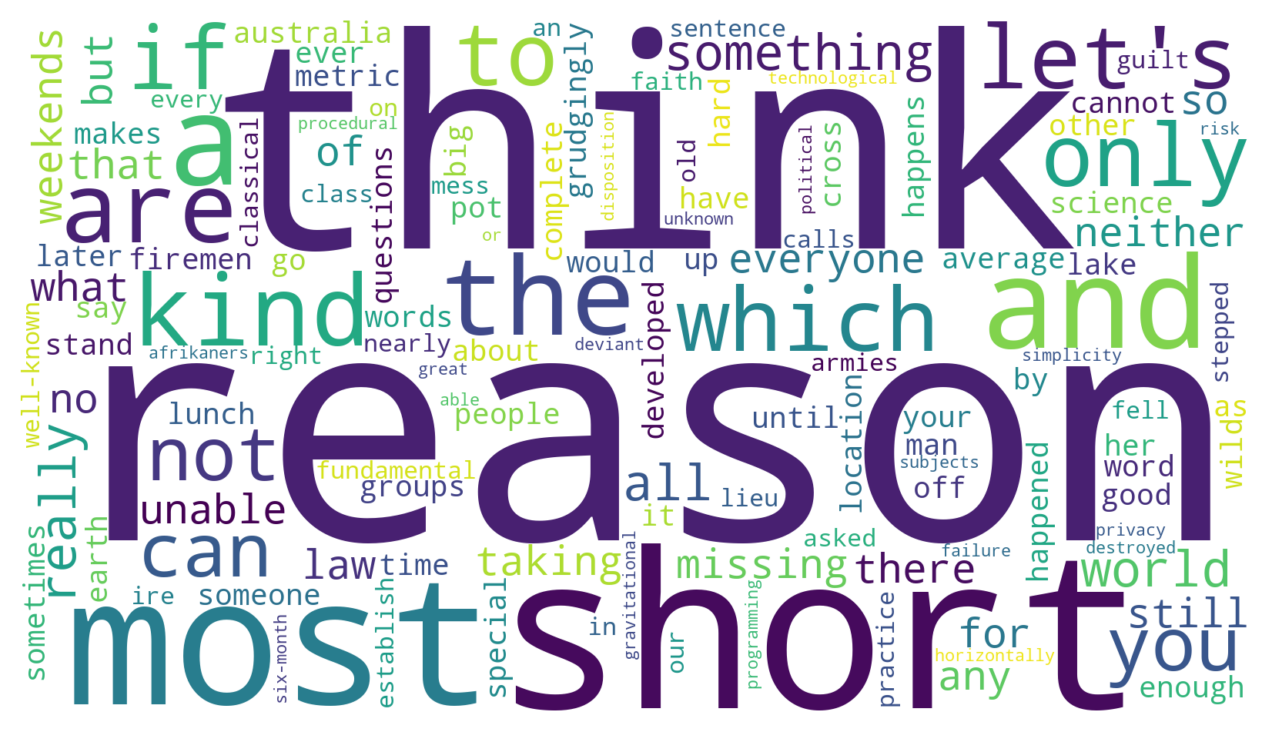}\\
{\small (d) Gemma-3-27B}
\end{minipage}
\caption{Word clouds of the words selected for suppression, pooled over the
five evaluation datasets. For readability the four most frequent
instruction-template tokens (\textit{sentences}, \textit{at}, \textit{then},
\textit{final}) are removed before rendering; other template tokens remain
visible (e.g., \textit{concise} and \textit{answer} for Qwen-3.5-9B;
\textit{reason}, \textit{think}, and \textit{short} for Gemma-3-27B).}
\label{fig:wordclouds}
\end{figure*}

Figure~\ref{fig:wordclouds} shows, for each model, a word cloud of the words
selected for suppression across the five evaluation datasets (pooled). The
suppressed words are predominantly function and discourse words---
connectives, pronouns, determiners, and negations---with a small tail of
named entities from individual questions.

\section{Cross-Model Judge Agreement}
\label{sec:cross_judge}

To assess whether the Self-Judge's verdicts reflect genuine dependence
judgments rather than model-specific biases, we re-judge sampled pivot words
from GSM8K with two external models---GPT-5.6 and
Gemini-3.1-Flash-Lite---and with human annotators, all given the identical
question. Cohen's $\kappa$ is computed pairwise between every pair of judges
for each generator model (Table~\ref{tab:cross_judge_agreement}).

\begin{table*}[h]
\centering
\small
\setlength{\tabcolsep}{5pt}
\begin{tabular}{l c c c c c c c}
\toprule
\textbf{Generator} &
\textbf{Self--GPT} &
\textbf{Self--Gemini} &
\textbf{Self--Human} &
\textbf{GPT--Gemini} &
\textbf{GPT--Human} &
\textbf{Gemini--Human} &
\textbf{Avg.\ $\kappa$} \\
\midrule
Qwen-3.5-4B   & 0.61 & 0.57 & 0.60 & 0.69 & 0.63 & 0.62 & 0.62 \\
Qwen-3.5-9B   & 0.66 & 0.63 & 0.65 & 0.72 & 0.67 & 0.66 & 0.67 \\
Gemma-3-12B   & 0.58 & 0.55 & 0.57 & 0.68 & 0.61 & 0.60 & 0.60 \\
Gemma-3-27B   & 0.64 & 0.61 & 0.63 & 0.71 & 0.66 & 0.65 & 0.65 \\
\bottomrule
\end{tabular}
\caption{Cross-judge agreement on PCI judgments on GSM8K.
For each generator model, pivot words are labeled by four judges: the
generator itself (Self), GPT-5.6, Gemini-3.1-Flash-Lite, and human
annotators (by the Authors). Cohen's $\kappa$ ($\geq 0.6$ indicates substantial agreement) is
computed pairwise over sampled pivot words.}
\label{tab:cross_judge_agreement}
\end{table*}

All judge pairs achieve $\kappa \geq 0.55$, and the Self-Judge agrees
substantially with the external models ($\kappa = 0.55$--$0.66$) and with
human annotators ($\kappa = 0.57$--$0.65$).
This cross-judge consistency supports the validity of using the generator
itself as a judge and suggests that the load-bearing/inert labels capture a
stable notion of dependence rather than idiosyncratic model preferences.

\subsection{Replacing the Self-Judge with External Judges}
\label{app:external_judge}
The $\kappa$ values in Table~\ref{tab:cross_judge_agreement} are computed
per word, over all candidate words whose removal flips a prediction, and
therefore measure chance-corrected agreement on whether each candidate is
load-bearing. A complementary question is whether this imperfect per-word
agreement matters for the intervention. We therefore replace the Self-Judge
in the Online KV Suppression pipeline with an external judge (GPT-5.6 or
Gemini-3.1-Flash-Lite), keep every other step unchanged, and measure
(i) downstream accuracy on GSM8K and (ii) Cohen's $\kappa$ between the
external judge and the Self-Judge on the \emph{selected} word, i.e., whether
the two pipelines suppress the same word (Table~\ref{tab:external_judge}).

\begin{table}[h]
\centering
\small
\setlength{\tabcolsep}{4pt}
\resizebox{\columnwidth}{!}{%
\begin{tabular}{lccccc}
\toprule
\textbf{Generator} & \textbf{Zero-Shot} & \textbf{Self} & \textbf{GPT} & \textbf{Gemini} &
\textbf{Selected-word $\kappa$ (GPT\,/\,Gemini)} \\
\midrule
Qwen-3.5-4B & 72.3 & 81.2 & 82.7 & 83.0 & 0.83\,/\,0.85 \\
Gemma-3-12B & 86.5 & 87.8 & 87.8 & 87.7 & 0.83\,/\,0.77 \\
\bottomrule
\end{tabular}}
\caption{GSM8K accuracy (\%) of Online KV Suppression when the Self-Judge is
replaced by an external judge, and Cohen's $\kappa$ between each external
judge and the Self-Judge on the selected word.}
\label{tab:external_judge}
\end{table}

External judges raise Qwen-3.5-4B accuracy by a further $1.5$--$1.8$~pp over
the Self-Judge and change Gemma-3-12B accuracy by at most $0.1$~pp, and the
selected-word $\kappa$ is high ($0.77$--$0.85$). Imperfect word-level
agreement therefore does not preclude effective intervention: the pipeline
is not tied to the generator's own verdicts, and a stronger external judge
can be substituted where one is available.

\section{Random and Low-Influence Inert Controls}
\label{app:inert_controls}
Table~\ref{tab:ablation_yes_no} compares suppressing the highest-influence
inert word (Ours) against influence-only, random-candidate, and
load-bearing selection. To test whether the \emph{ranking} among inert words
matters, we add two controls that share the judge verdicts and suppression
operator of Ours but change only which inert word is suppressed:
\textbf{R} suppresses one randomly chosen inert word other than the
top-ranked one, and \textbf{L} suppresses the inert word with the
\emph{lowest} $\Delta_k$. Table~\ref{tab:inert_controls} reports accuracy on
GSM8K and AQUA-RAT.

\begin{table}[h]
\centering
\small
\setlength{\tabcolsep}{4pt}
\resizebox{\columnwidth}{!}{%
\begin{tabular}{llcccc}
\toprule
\textbf{Model} & \textbf{Dataset} & \textbf{Zero-Shot} & \textbf{Ours} & \textbf{R} & \textbf{L} \\
\midrule
\multirow{2}{*}{Qwen-3.5-4B} & GSM8K    & 72.3 & \textbf{81.2} & 75.6 & 74.5 \\
                             & AQUA-RAT & 69.7 & \textbf{73.2} & 72.4 & 70.5 \\
\midrule
\multirow{2}{*}{Gemma-3-12B} & GSM8K    & 86.5 & \textbf{87.8} & 87.0 & 86.1 \\
                             & AQUA-RAT & 65.4 & \textbf{67.3} & 65.0 & 65.7 \\
\midrule
\multicolumn{2}{l}{Mean gain over Zero-Shot} & -- & $+3.9$ & $+1.5$ & $+0.7$ \\
\bottomrule
\end{tabular}}
\caption{Accuracy (\%) when suppressing the highest-influence inert word
(Ours), a random non-top inert word (R), or the lowest-influence inert word
(L). All three use the same Self-Judge verdicts and KV suppression operator.}
\label{tab:inert_controls}
\end{table}

Ours exceeds both controls in all four settings, with a mean gain of
$+3.9$~pp over zero-shot versus $+1.5$~pp for R and $+0.7$~pp for L.
Suppressing an arbitrary inert word thus yields only a small and
inconsistent benefit; the gain of Ours comes from selecting the inert word
with the largest measured influence, confirming that influence-based ranking
carries signal beyond the judge's binary verdict.

\section{From Token Flips to Response Changes}
\label{app:local_vs_response}
A single flipped token need not change the meaning of a response, so it is
natural to ask how suppressing one prompt word produces the accuracy gains
in Table~\ref{tab:main_results}. We compare the zero-shot and
post-suppression responses of Qwen-3.5-4B on the GSM8K questions where an
inert word was found and suppressed, at three levels
(Table~\ref{tab:local_vs_response}). \textbf{L1} classifies the first word
at which the two responses diverge as numeric, logical, or entity-related
versus an ordinary word. \textbf{L2} asks an LLM judge whether the chain of
thought changed substantively (a different intermediate quantity, operation,
or claim) rather than being paraphrased. \textbf{L3} records whether the
final answer changed, and in which direction.

\begin{table}[h]
\centering
\small
\setlength{\tabcolsep}{5pt}
\resizebox{\columnwidth}{!}{%
\begin{tabular}{llc}
\toprule
\textbf{Level} & \textbf{Metric} & \textbf{\%} \\
\midrule
\multirow{2}{*}{L1} & First divergence: numeric / logic / entity & 45.3 \\
                    & First divergence: ordinary word            & 54.7 \\
\midrule
L2 & Substantive CoT change (LLM judge) & 29.4 \\
\midrule
\multirow{3}{*}{L3} & Final answer changed      & 23.0 \\
                    & \quad wrong $\to$ right   & 13.5 \\
                    & \quad right $\to$ wrong   & 4.6 \\
\bottomrule
\end{tabular}}
\caption{How a single suppressed prompt word propagates through the response
(Qwen-3.5-4B, GSM8K, intervened questions only).}
\label{tab:local_vs_response}
\end{table}

Nearly half of the first divergences involve numbers, logic, or entities,
which can redirect a multi-step computation. Because the KV entries remain
zeroed for the entire redecode, such a divergence is not corrected by the
surrounding context but propagates: $29.4\%$ of intervened responses show a
substantive CoT change and $23.0\%$ reach a different final answer.
Corrections outweigh regressions ($13.5\%$ vs.\ $4.6\%$, a net $+8.9$~pp),
so the downstream changes are beneficial on balance. A local flip is thus
the entry point of a different reasoning trajectory rather than an isolated
lexical change.

\section{Detection Operator: Deletion vs.\ KV Suppression}
\label{app:detection_operator}
Detection (\S\ref{sec:loo_resp}) scores each prompt word by textually
deleting it, whereas the intervention (\S\ref{sec:method}) zeroes the word's
KV-cache entries while keeping the token in place. We use deletion for
detection because it matches the counterfactual the Self-Judge is asked
about, and position-preserving suppression for the intervention so that the
model still answers the original, intact question. The two operators are
nevertheless not identical, so we re-run detection with KV suppression as
the operator: $\Delta_k$ is computed from the response log-probability under
the KV-suppressed prompt instead of the deleted prompt, with the rest of the
pipeline unchanged (Table~\ref{tab:detection_operator}).

\begin{table}[h]
\centering
\small
\setlength{\tabcolsep}{5pt}
\begin{tabular}{lccc}
\toprule
\textbf{Generator} & \textbf{Deletion} & \textbf{KV-supp.} & \textbf{Agreement} \\
\midrule
Qwen-3.5-4B & 81.2 & 80.9 & 90.3\% \\
Gemma-3-12B & 87.8 & 88.0 & 92.0\% \\
\bottomrule
\end{tabular}
\caption{GSM8K accuracy (\%) of Online KV Suppression with deletion (default)
vs.\ KV suppression as the detection operator, and the fraction of questions
on which both operators select the same word.}
\label{tab:detection_operator}
\end{table}

Accuracy changes by $-0.3$~pp (Qwen-3.5-4B) and $+0.2$~pp (Gemma-3-12B), and
the two operators select the same word on over $90\%$ of questions. The
choice of detection operator therefore has little effect on which words are
suppressed or on final accuracy.

\section{Latency}
\label{app:latency}
Online KV Suppression runs four stages: the initial greedy generation, one
batched leave-one-out scoring pass over the prompt words, Self-Judge queries
in descending order of $\Delta_k$ with early stopping at the first inert
verdict, and a single redecode. Table~\ref{tab:latency} reports the
wall-clock time of each stage on GSM8K, measured by running the pipeline of
the main experiments stage by stage on five questions per model.

\begin{table}[h]
\centering
\small
\setlength{\tabcolsep}{3.5pt}
\resizebox{\columnwidth}{!}{%
\begin{tabular}{lcccccccc}
\toprule
\textbf{Generator} & \textbf{Generate} & \textbf{LOO} & \textbf{Judge} & \textbf{Redecode} & \textbf{Total} & \textbf{Supp.} & \textbf{Expected} & \textbf{$\times$ZS} \\
\midrule
Qwen-3.5-4B & 3.4  & 1.9  & 0.3 & 3.4  & 9.1  & 92.1\% & 8.8  & 2.6 \\
Gemma-3-12B & 19.3 & 13.6 & 0.5 & 19.7 & 53.1 & 15.9\% & 36.6 & 1.9 \\
\bottomrule
\end{tabular}}
\caption{Per-question wall-clock latency (seconds) of each stage of Online KV
Suppression on GSM8K, averaged over five questions per model on one NVIDIA
A100 80GB PCIe GPU (target model in HF Transformers with eager attention,
Self-Judge served by vLLM).
Redecode is the cost when an inert word is found and Total assumes all four
stages run. Supp.\ is the fraction of GSM8K questions on which an inert word
is found in the main experiments, Expected is
Generate\,+\,LOO\,+\,Judge\,+\,Supp.$\times$Redecode, and $\times$ZS is
Expected relative to a zero-shot decode (Generate).}
\label{tab:latency}
\end{table}

Generation and redecoding dominate, each costing about one zero-shot decode.
The batched leave-one-out pass costs roughly half a decode for Qwen-3.5-4B and
0.7 of a decode for Gemma-3-12B, whose 262k-token vocabulary makes the
teacher-forced log-softmax expensive. The Self-Judge, which stops after
$1.2$--$1.8$ queries on average, adds under half a second. Because redecoding
runs only when an inert word is found ($92.1\%$ of GSM8K questions for
Qwen-3.5-4B and $15.9\%$ for Gemma-3-12B in the main experiments), the
expected end-to-end latency is $8.8$~s and $36.6$~s per question, i.e.,
$2.6\times$ and $1.9\times$ a zero-shot decode. Gemma-3-12B responses run to
the 350-token budget, so its decode stages are correspondingly longer.
Absolute times reflect unoptimized single-question decoding; the ratios are
the relevant quantity.

\section{Case Study}
\label{sec:casestudy}

\begin{figure}[t]
  \centering
  \includegraphics[width=\columnwidth]{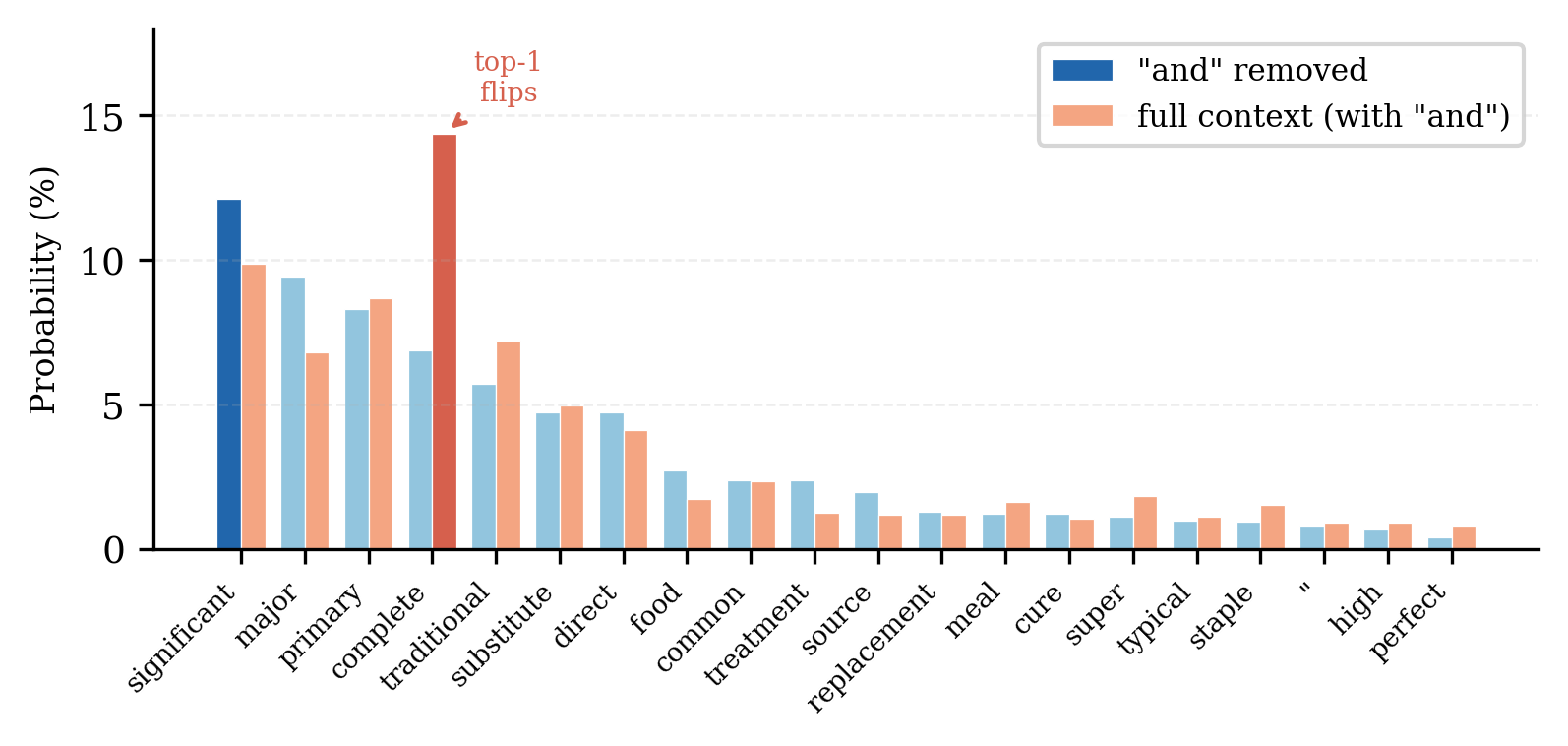}
  \caption{Predicted token distribution with and
  without the connector \textit{and} (TruthfulQA, Qwen-3.5-4B). Red: full
  context including \textit{and} (top-1 = \textit{complete}, 14.4\%). Blue:
  \textit{and} removed from the otherwise intact context (top-1 =
  \textit{significant}, 12.1\%).}
  \label{fig:case_barplot}
\end{figure}

Figure~\ref{fig:case_barplot} illustrates the contradiction introduced in
Section~\ref{sec:introduction}. The target position arises
in the partial CoT
``\ldots\textit{completely safe and harmless. While they are not a}~[?]''
(TruthfulQA, Qwen-3.5-4B). Under the full context,
\textit{complete} is the top prediction (14.4\%). Removing the coordinator
\textit{and}, located 17 words before the target position, while keeping the
rest of the context intact shifts the top prediction to
\textit{significant} (12.1\%). The Self-Judge nevertheless classifies
\textit{and} as irrelevant. This example captures the core PCI pattern:
a semantically weak word exerts a measurable effect on generation that the
model's own judgment does not endorse.

\begin{algorithm}[t]
\caption{Online KV Suppression}
\label{alg:suppress}
\begin{algorithmic}[1]
\Require Prompt $x$ (words $W_1,\ldots,W_N$), model $\mathcal{M}$,
         judge $\mathcal{J}$, max tokens $T_{\max}$
\State $\mathcal{R} \leftarrow \operatorname{greedy}(\mathcal{M},x,T_{\max})$
       \Comment{generate response}
\State compute $\Delta_k\;\forall k$ over $\mathcal{R}$
       (Eqs.~\ref{eq:chunk_logprob}--\ref{eq:chunk_adjk})
       \Comment{one batched pass}
\State $\mathcal{T} \leftarrow W_1,\ldots,W_N$ sorted by $\Delta_k$
       descending;\; $w^\star \leftarrow \bot$
\For{$W_k \in \mathcal{T}$} \Comment{judge top-down}
  \If{$\mathcal{J}(W_k,\mathcal{R}) = \textsc{no}$}
       $\;w^\star \leftarrow W_k$;\; \textbf{break}
  \EndIf
\EndFor
\If{$w^\star = \bot$} \Return $\mathcal{R}$ \Comment{no spurious word} \EndIf
\State $\mathcal{K} \leftarrow \mathcal{M}(x)$;\;
       zero KV at $\mathrm{pos}(w^\star)$ \Comment{rebuild \& suppress}
\State \Return $\operatorname{greedy}(\mathcal{M},\mathcal{K},T_{\max})$
       \Comment{redecode response}
\end{algorithmic}
\end{algorithm}

\end{document}